# Do media narratives shape foreign ownership? Evidence from African regions

Andrea Fronzetti Colladon[a,*,1], Roberto Vestrelli[b,c,1], Stefania Bait[d,e], Massimiliano Maria Schiraldi[e]

[a] Department of Civil, Computer Science and Aeronautical Technologies Engineering, Roma Tre University, Via della Vasca Navale 79, Rome 00146, Italy
[b] Department of Engineering, University of Perugia, Via G. Duranti 93, Perugia 06125, Italy
[c] Sloan School of Management, Massachusetts Institute of Technology, 100 Main St, Cambridge, MA, 02142, USA
[d] United Nations Industrial Development Organization (UNIDO), Agro-Industries and Skills Development Division, Wagramer Str. 5, Wien 1220, Austria
[e] Department of Enterprise Engineering, University of Rome Tor Vergata, Via del Politecnico 1, Rome 00133, Italy

ABSTRACT

Foreign investors rarely observe local business conditions directly, especially in developing economies where reliable subnational information is limited. In such contexts, media narratives may shape how investment locations are perceived by increasing their visibility and framing local risks and opportunities. This paper examines the relationship between media narratives and foreign ownership across African locations. We combine firm-level data from the World Bank Enterprise Surveys with a novel corpus of 167,746 English-language news articles on industrial investment in Africa published between 2014 and 2019. Using text mining and semantic network analysis, we construct investment-location measures of the semantic importance of business-related media themes through the Semantic Brand Score (SBS) and study their associations with firm-level foreign ownership. The results point to a significant association between how the media frames a location and the level of foreign presence there. This association varies across frames. It is predominantly positive, relates to both the intensive and extensive dimensions of foreign ownership depending on the theme, and may interact negatively with the investment climate. This paper contributes to research on the geography of foreign capital by identifying public narratives as a feature of the information environment associated with observed foreign-equity participation.

## 1. Introduction

Why is foreign capital more prevalent in some African locations than in others? A large literature emphasizes the role of macroeconomic fundamentals, institutional quality, infrastructure, human capital, and political stability in shaping foreign direct investment

* Correspondence to: Roma Tre University – Department of Civil, Computer Science and Aeronautical Technologies Engineering, Via della Vasca Navale 79, Rome 00146, Italy.

*E-mail addresses:* andrea.fronzetticolladon@uniroma3.it (A. Fronzetti Colladon), roberto.vestrelli@dottorandi.unipg.it (R. Vestrelli), stefania.bait@uniroma2.it (S. Bait), max.schiraldi@uniroma2.eu (M.M. Schiraldi).

[1] The first two authors contributed equally to this work and are listed in alphabetical order

(FDI). Yet foreign investors rarely observe these conditions directly and completely. Instead, their perceptions of places are partly mediated by publicly available information, including news media narratives. This paper examines whether and how media coverage of subnational locations in Africa is associated with the presence and intensity of foreign ownership. We show that foreign ownership is related not only to the amount of media coverage a location receives, but also to the way that location is framed in the news.[2]

Industrial development has long been a policy priority in Africa because of its potential to create employment and support broader economic transformation (Samouel and Aram, 2016). Yet progress across the continent has been slow and highly uneven. Accelerating this process remains a stated goal of both the UN Agenda 2030 and the African Union Agenda 2063 (African Union Commission, 2015; Kamalam, 2017; UNIDO, 2020). Foreign investment is one possible source of the capital and knowledge required for this transformation, particularly when it helps expand productive capacity and connects local economies to international markets. These benefits, however, are not automatic. They depend on where foreign capital is invested and on the extent to which foreign firms become connected to the local economy. Previous research shows that spillovers are stronger when local institutions and policy frameworks facilitate interactions between foreign firms and domestic actors (Crespo and Fontoura, 2007; Amendolagine et al., 2013; Du, Harrison and Jefferson, 2014; Zhang et al., 2021). The African pattern illustrates this point. In 2018, FDI inflows rose by 11 % to $46 billion, but UNCTAD attributes much of that increase to continued resource-seeking investment. Greenfield investment in manufacturing also rose to $33 billion, yet projects in industries associated with early industrialization remained weak, especially in low-income African countries (UNCTAD, 2019a).

A substantial body of research has examined the macroeconomic and institutional barriers that limit foreign capital inflows. Attracting foreign investment requires a wide range of complementary inputs, including skilled labor, reliable infrastructure, functioning public services, access to finance, and predictable institutions (Cleeve et al., 2015). Corruption, weak trade openness, financial constraints, limited information technology, poor-quality public services, political instability, and skills shortages have all been identified as factors that discourage foreign investors (Asongu and Odhiambo, 2020; Bahoo et al., 2020; Kinda, 2010; Asongu, 2020; Calabrese and Tang, 2023; Dupasquier and Osakwe, 2006; Okafor et al., 2017; Saini and Singhania, 2018; Wyrwa, 2019).

These barriers ultimately enter firms' decisions over whether to invest in a location and, conditional on investment, how large an ownership stake to hold. Both decisions require investors to assess potential host locations, and this assessment depends in part on the information they collect before committing capital. Matějka and McKay (2015) show that when alternatives cannot be fully investigated, choices depend on which information is acquired. Van Nieuwerburgh and Veldkamp (2009) show that information affects not only which assets investors choose, but also the size of the positions they are willing to take. News can form part of this information set, but its role is not limited to determining the extent of media coverage for a given location. The content of that coverage can also affect which aspects of a location become most prominent in investors' assessments. Salience theory predicts that attributes that stand out in the available information can receive greater weight when alternatives are evaluated (Bordalo et al., 2013). Repeated coverage of corruption, for example, may make corruption a more prominent consideration when a location is assessed, whereas other reporting may direct attention toward infrastructure, labor conditions, or other features of the local business environment. Existing research provides little evidence on whether the themes through which a location is framed in the news are related to foreign ownership.

To address this gap, we combine text mining with social network analysis following the semantic network approach proposed by Fronzetti Colladon (2018). This method allows us to quantify not only how much media coverage a location receives, but also how that location is framed by measuring the semantic importance of specific frames in the surrounding media discourse. We apply this approach to a novel dataset of approximately 167,000 news articles on industrial development in Africa. We extract location-specific media coverage and measure the semantic importance of a wide set of business-related themes associated with each location.[3] We then link these media-derived measures to firm-level data from the World Bank Enterprise Surveys (WBES) to examine the association between media frames and foreign ownership at the subnational level.[4]

Africa provides a particularly relevant setting for this analysis. For this continent, foreign investors often have to evaluate locations where reliable public information is limited. Statistical systems remain uneven across the continent, particularly at subnational levels, leaving substantial gaps in what can be learned about individual places from official sources (Chen and Nordhaus, 2011; Jerven and Johnston, 2015). At the same time, foreign investment is highly concentrated in a relatively small number of countries, regions, and activities. Under these conditions, publicly available news becomes one of the channels through which less familiar locations enter the information set of foreign investors (Kulchina, 2014). How these locations are portrayed in the news can therefore be especially relevant when investors have little direct information to base their assessment on (Refakar and Gueyie, 2020; Wilke et al., 2024; Ackermann et al., 2025).

[2] Throughout this paper, when we use the term "frame," we primarily refer to the salience of a particular aspect of reality that news outlets emphasize when reporting on a given location. This definition of "frame" is based on the formulation proposed by Entman (1993). This idea will be explained in greater detail in Section 2.1.

[3] The themes analyzed in the news were selected a priori based on the list of business obstacles identified in the World Bank Enterprise Surveys (WBES), as these reflect key challenges that firms commonly report. Specifically, we focus on the following themes: access to land (g30a), access to finance (k30), electricity (c30a), transport (d30a), customs and trade regulations (d30b), crime, theft and disorder (i30), courts (h30), taxation (j30a and j30b, grouped under a single "taxes" theme), business licensing and permits (j30c), political instability (j30e), corruption (j30f), labor regulations (l30a), and inadequately educated workforce (l30b). The codes in parentheses refer to the obstacle identifiers used in the WBES questionnaire.

[4] Because the WBES outcome measures each firm's foreign-equity share at the survey date, our empirical analysis concerns the observed presence and intensity of foreign capital, not new FDI flows, entry timing, or a directly observed location choice.

We report two main findings. First, foreign ownership is associated not only with the amount of news coverage a location receives, but also with the semantic centrality of specific themes within media narratives about that location. Among the thirteen business themes considered, three are associated with foreign ownership: access to land, inadequately educated workforce, and crime, theft and disorder. In the contemporaneous baseline specification, greater semantic centrality of land- and workforce-related themes is associated with higher foreign ownership, whereas the coefficient of crime, theft and disorder is negative. However, the latter relationship is not stable across specifications: the coefficient becomes positive when media measures are lagged and when we use an instrumental-variable strategy. Although this warrants further research, the overall results lead us to conclude that the direct relationship between media content and foreign ownership is generally positive. These associations operate along different margins: access to land is primarily associated with the intensive margin of foreign ownership, whereas the other two themes are primarily associated with the extensive margin.[5]

The second result we find is that certain media frames may moderate (negatively) the relationship between foreign ownership and variables that measure the investment climate in a location. In locations where the media place significant emphasis on issues related to the judicial system (e.g., legal infrastructure, judicial reforms, and access to justice), obstacles related to infrastructure and human capital are negatively associated with foreign ownership. This negative association, however, is not observed in regions where the media do not emphasize such issues.

This study makes three main contributions. First, it adds an information dimension to the study of foreign ownership. Existing research has largely focused on the economic and institutional characteristics that make locations more or less attractive to foreign firms. We examine a different part of the investment process: the information through which investors can learn about those characteristics. By linking foreign equity to the news coverage of individual locations, we show how the local information environment can be studied alongside conventional determinants of foreign investment. Second, the paper identifies which aspects of that information environment are most closely associated with foreign ownership. The results show that the relevance of news coverage depends on the issues that become semantically central around a location, rather than simply on how frequently that location appears in the news. This distinction is particularly important where investors face substantial information frictions and where public and private organizations devote resources to providing information about local investment conditions. Third, we provide a way to systematically measure these location-specific information patterns. Using semantic networks and the Semantic Brand Score, we quantify the semantic centrality of business-related themes in the news surrounding each location and link these measures to firm-level foreign ownership. This makes it possible to incorporate information derived from public news into empirical analyses of local investment environments.

## 2. Literature review

This paper builds on the idea that foreign investors evaluate locations not only through objective economic and institutional indicators but also through the information environments in which those locations are framed. This is particularly relevant in developing economies, where reliable local information is often scarce, fragmented, or costly to acquire. In such contexts, the media may affect foreign ownership in two related ways. First, media can increase the visibility of specific locations and reduce information asymmetries. Second, media can frame locations by making some issues more salient than others, thereby shaping how investors interpret local risks and opportunities.

The literature review develops these arguments in two steps. Section 2.1 reviews the literature on media and foreign investments. This literature motivates our first empirical analysis, which examines whether foreign ownership is associated not only with the amount of media coverage a location receives but also with the specific themes through which it is framed.

Section 2.2 then turns to the literature on foreign investment determinants with a specific focus on the investment climate. Since prior research has shown that obstacles that shape the investment climate in a location (such as taxation, infrastructure, crime, corruption, courts, and labor-market constraints) affect foreign ownership, we examine whether media portrayals of these obstacles moderate that relationship.

### 2.1. Media and FDI

The role of the media in the capital market is increasingly studied in the international business literature. By reaching a broad population of investors, mass media can alleviate informational frictions by reducing information asymmetries between capital providers and recipients and influence security prices, even without providing genuine news (Fang and Peress, 2009; Bonsall, Green and Muller, 2020; Gao et al., 2024). International evidence shows that media coverage is associated with firms' investment decisions, with information asymmetry and corporate governance acting as key channels (Gao et al., 2025). The media can also serve as an informal governance institution by exerting reputational pressure on firms, governments, and public officials, while raising the probability of scrutiny by the public and administrative authorities (Dai et al., 2015; Zhou et al., 2016; Chen et al., 2021).

The role of the media is also investigated in relation to foreign investments. Empirically, Kulchina (2014) shows that greater foreign-media coverage of Russian cities increases subsequent foreign entry, with stronger effects for firms with less private information about local markets, suggesting therefore that media works by reducing information asymmetry. A similar empirical pattern can also be observed at the country level. Adams et al. (2018) find that greater coverage of developing countries in The Economist is associated with larger FDI flows from U.S. multinationals. Li et al. (2021) show that greater digital-media freedom in host countries increases the likelihood that emerging-market multinationals enter through wholly owned subsidiaries.

[5] For clarity, we refer to the extensive margin as whether a firm is foreign-owned or not, using a binary dependent variable equal to one when foreign ownership exceeds 10% and zero otherwise. We refer to the intensive margin as the degree of foreign ownership, using the continuous percentage of foreign ownership as the dependent variable among firms with positive foreign ownership.

However, greater media visibility of a location does not directly translate to a larger presence of foreign capital. While greater coverage may increase awareness of a location and reduce the costs of acquiring information, the content of that coverage may also cause concerns about risk, instability, or weak governance. Refakar and Gueyie (2020) find that more corruption stories about a host country in The Wall Street Journal are associated with lower U.S. FDI flows to that country. Recent evidence on international portfolio allocation shows that hostile media sentiment toward a country predicts lower cross-border portfolio exposure to that country, suggesting that negative media narratives can shape investors' risk perceptions and international capital allocation decisions (Li et al., 2025).

This double effect of the media raises a question. How are location media frames associated with foreign ownership? Two locations receiving similar amounts of news coverage can be represented through very different frames. One location may appear frequently in articles discussing crime and security, while another may appear in connection with land, workforce conditions, courts, or infrastructure. Our first analysis adds to the literature about media and foreign investments by examining how and which media frames of a location are associated with foreign ownership. This approach recognizes that foreign investors may respond not only to how often a location is mentioned in the news, but also to what is said about that location and to the importance of particular themes within media narratives. In doing so, we extend existing work on media and foreign ownership by introducing a measure of location-specific media framing and showing its association with the presence of foreign ownership.

We draw on two main theories that we use to (1) motivate our methodological choice of the SBS, and (2) guide our empirical exploration. The first is Entman's (1993) framing theory. According to Entman (1993), framing involves selecting "some aspects of a perceived reality and make them more salient in a communicating text". In this sense, the media can represent otherwise comparable locations differently by emphasizing different aspects of their business environment. We use the Semantic Brand Score (SBS) to operationalize the "salient" component of frame as proposed by Entman. Originally developed to measure brand importance, SBS also applies to sets of words representing broader concepts. It combines prevalence (frequency of mention), diversity (the variety and uniqueness of associated words), and connectivity (the extent to which a concept links other words in a co-occurrence network) (Fronzetti Colladon, 2018) to produce a measure of the semantic importance of a concept in a text. Drawing on associative network-based memory models, Toschi, Torrisi, and Fronzetti Colladon (2025) demonstrate that memorability stems from the combined contribution of these three components. Applied to news about a location, we use the SBS to measure the "salience", to use Entman (1993) words, of "some aspects of a perceived reality", and translate them into quantitative measures.

Entman (1993) further argues that increasing the salience of a concept makes it more likely to be noticed by receivers. This provides a natural link to the salience theory of Bordalo et al. (2013), according to which attention is disproportionately directed toward salient elements of a choice problem. They show that decision-makers do not fully weigh all available information equally but instead overweight the frames that capture their attention. This generates a context-dependent representation of the alternatives under consideration. We argue that the aspects of a location made salient through media framing may therefore receive disproportionate weight when investors evaluate investment locations. Consequently, different media frames may lead investors to place different weights on the aspects associated with a location and, in turn, differently affect the presence and intensity of foreign ownership.[6]

Africa provides a particularly relevant setting for our analysis. The continent is characterized by persistent information asymmetries, limited and uneven statistical coverage, and relatively low levels of foreign investment (Chen and Nordhaus, 2011; Jerven and Johnston, 2015; Ackermann et al., 2025). In such contexts, investors may rely more heavily on public narratives to compensate for the lack of reliable and comparable local information.[7] Media representations may therefore have stronger associations with how investment locations are perceived. Large-scale content analyses suggest that negative or stereotypical portrayals of African countries can increase perceived risks and raise sovereign borrowing costs substantially (Wilke et al., 2024). Studying Africa, therefore, addresses an important development issue while also providing a useful empirical context for examining how media narratives mediate information frictions in global capital allocation.

### 2.2. Foreign Ownership Determinants

A large literature examines why some countries and regions attract more foreign investment than others. Early and contemporary studies have identified several macroeconomic determinants of foreign ownership, including exchange-rate conditions, inflation, labor costs, market size, skills availability, and corporate taxation (Cheng and Kwan, 2000; Chakrabarti, 2001; Asiedu, 2002). These factors influence expected profitability and shape the relative attractiveness of alternative host locations. Institutional quality is also central. The bureaucratic, judicial, and political environment of a host country can substantially affect investment decisions (Bhupatiraju,

[6] From a theoretical perspective, the frames through which a location is portrayed can affect both the extensive and intensive dimensions of foreign ownership. Also, the direction of this association is ambiguous. Some frames may favor the presence of foreign ownership, while others may discourage it. Because this study's nature is mainly empirical, we do not theoretically model whether the semantic importance of these frames might occur with one dimension rather than the other, nor their direction. We leave this theoretical exploration as an avenue for future research. In our empirical analysis, we take both possibilities into account.

[7] It is important to clarify that all the media metrics developed and discussed throughout the paper are not intended to serve as a proxy for investors' readership. As explained in the previous paragraph, our focus is on the framing effect, regardless of readership size.

2020; Kinda, 2010; Krifa-Schneider et al., 2022). Weak property-rights protection may deter multinational enterprises by increasing the risk of expropriation or appropriation (Sharma and Bandara, 2010; Seyoum, 2011). Excessive regulation can discourage entry by raising the costs of establishing and operating a business (Naudé and Krugell, 2007), while bureaucracy and corruption can increase uncertainty and transaction costs (Anghel, 2005). Recent evidence further shows that economic policy uncertainty discourages foreign inflows, whereas stronger institutional quality directly attracts foreign investment and mitigates the adverse effect of policy uncertainty (Tabash, 2025).

More recently, the World Bank Enterprise Surveys have allowed researchers to examine these issues at the firm level across a broad set of developing countries, including many African economies (Kinda, 2010; Bhupatiraju, 2020). This literature focuses on the investment climate, usually measured through the business obstacles perceived by firms. These obstacles include taxation, regulation, access to finance, infrastructure, corruption, crime, courts, and labor-market constraints. Because these measures reflect how firms experience their operating environment, they provide an important window into the constraints that may shape foreign investment decisions.

Empirical evidence confirms that perceived business obstacles are associated with foreign ownership, although the direction and strength of these associations vary across contexts and types of constraints. Gelb et al. (2008), for example, show that firm-reported obstacles are correlated with broader institutional features of the countries in which firms operate. Kinda (2010) finds that perceived constraints related to physical and financial infrastructure discourage foreign ownership. Bhupatiraju (2020) shows that perceived problems with internet availability and shortages of skilled labor negatively affect foreign investment decisions. At the same time, some infrastructure constraints, such as electricity and transportation, may be associated with higher inward FDI. In contrast, institutional obstacles, such as corruption and crime, are not always statistically significant determinants. Overall, this literature shows that investment climate variables are relevant to understanding foreign ownership, but their effects are neither uniform nor mechanically predictable.

In this article, we examine whether the relationship between the investment climate and foreign ownership in a location is moderated by the way the media frame the location in light of these obstacles. From a theoretical perspective, the direction of this moderating effect is ambiguous. From an information intermediation perspective, greater media focus on a specific frame, such as corruption, can weaken the negative relationship between that business obstacle and foreign ownership. By disseminating information about the issue, media coverage can lower investors' information acquisition costs and increase investor knowledge. Investors may be better able to assess the obstacle and incorporate related information into their evaluation of the location. A negative moderating effect, on the other hand, can be explained by the salience theory described in Section 2.1. When a specific theme becomes highly central in the media representation of a location, foreign investors may assign greater weight to it. As a result, a business obstacle that would otherwise have limited influence on foreign ownership may become more relevant for investment decisions. In empirical terms, this implies that a weak relationship between a perceived business obstacle and foreign ownership may become more negative when media coverage places greater emphasis on that obstacle. Our empirical analysis contributes to the literature on the determinants of foreign ownership by examining whether the media moderates the relationship between the investment climate and foreign ownership, and by investigating whether such moderation occurs, which themes it involves, and in which direction.

## 3. Data collection and methodology

This section describes the data sources, variable construction, and empirical strategy used to examine the relationship between media narratives, local business conditions, and foreign ownership. The analysis combines two complementary sources of information. First, we use firm-level data from the World Bank Enterprise Surveys (WBES), which provides information on firms' ownership structure, perceived business obstacles, and firm characteristics. Second, we collect online news articles from Event Registry and use text mining and semantic network analysis to construct location-specific measures of media framing.

Fig. 1 summarizes the conceptual framework of the empirical analysis. The first relationship examined in the paper links media variables to foreign ownership. This analysis tests whether the semantic importance of specific business-related themes in media coverage is associated with the presence of foreign capital in a location. The second relationship introduces media content variables as moderators of the association between firms' perceived business obstacles and foreign ownership. This second analysis assesses whether media frames shape the way investment climate constraints are associated with foreign ownership.

Throughout the paper, the term "region" refers to the geographic areas identified by the a2x code in the WBES questionnaire. According to the WBES documentation, this code identifies "regions within a country that represent the largest centers of production and business enterprise." Therefore, any reference to a region in the remainder of the paper denotes these WBES-defined subnational areas. Table A3 in the Appendix summarizes all variables used in the empirical analysis, their data sources, and their level of measurement.

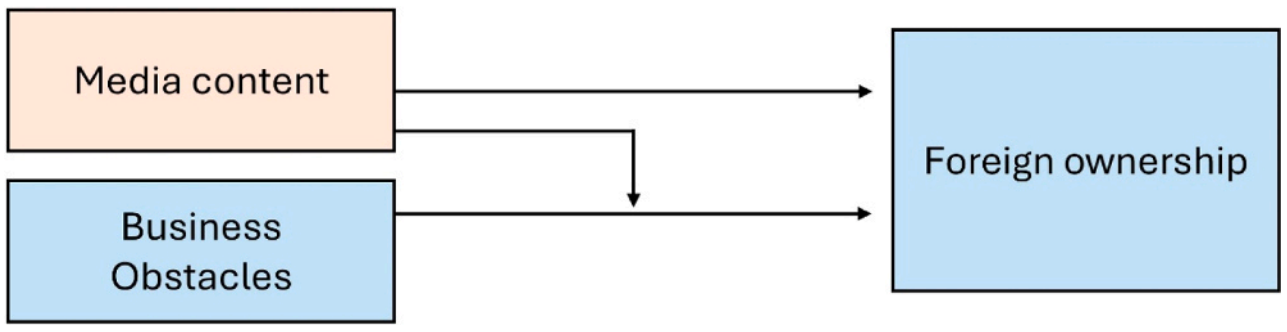

**Fig. 1.** Conceptual framework linking media content, business obstacles, and foreign ownership.

### 3.1. Firm-level data and investment climate measures

The WBES provides a rich and standardized source of firm-level data for studying foreign ownership and local business conditions. The survey collects harmonized cross-country information on firms' ownership structures, which allows us to measure foreign ownership directly at the firm-level. It also asks firms to evaluate a set of business obstacles that affect their daily operations. These firm-level perceptions provide a geographically and contextually grounded measure of the local investment climate, which is particularly useful in settings where aggregate institutional indicators may be unavailable or insufficiently granular.

The WBES is designed to collect qualitative and quantitative information on firms and the business environments in which they operate. The surveys are completed by firm managers and owners and cover topics such as infrastructure, trade, regulation, taxation, business licensing, corruption, crime, courts, finance, and labor. They also include information on firm characteristics, such as age, size, sector, and ownership. The WBES is conducted by the World Bank and its partners and covers small, medium-sized, and large firms across several regions worldwide. Although enterprise surveys have existed since the 1990s, the adoption of the Global Methodology from 2005 to 2006 onward improved comparability across countries. Surveys following this methodology are stratified by sector, location, and firm size, making the data representative within these strata. WBES data have been widely used to study firm productivity, foreign ownership, FDI spillovers, and investment climate constraints (Rutkowski, 2006; Héricourt and Poncet, 2009; Kinda, 2010; Bhupatiraju, 2020; Xu, Liu and Abdoh, 2022).

Our sample covers 29 African countries, 107 WBES regions, and 6232 firms surveyed between 2014 and 2019. Tables A4 and A5 in the Appendix report the countries and regions included in the sample, respectively. Summary statistics for all variables used in the analysis are provided in Table A3.

The dependent variable in all regressions is foreign ownership, measured as the percentage of the firm's capital owned by foreign private individuals, companies, or organizations at the survey date. The main firm-level controls include firm age, firm size, sales growth, and employment growth. Firm age is measured as the natural logarithm of the number of years the firm has been operating at the time of the survey. Firm size is a categorical variable distinguishing small firms, medium-sized firms, and large firms. This classification is provided by the WBES and has been widely adopted in previous studies (e.g., Kinda, 2010; Refakar and Gueyie, 2020). Employment growth and sales growth measure, respectively, the percentage growth in employment and sales relative to the last fiscal year. While WBES surveys collect data at a specific point in time, some questions also ask for fiscal data relative to the previous operating year, such as employment and sales, which allow us to construct growth rather than level variables.

We also include four location-level controls. The first is an agglomeration measure, defined as the number of foreign-owned firms operating in the same region-sector combination. This variable captures the local concentration of foreign capital and proxies for the attractiveness of a given region-sector environment to foreign investors. The second is region size, measured as the number of surveyed firms located in each WBES region. This variable controls for differences in the scale of local economic activity captured by the survey. The third is the median nighttime luminosity of a location, which we use as a proxy for the intensity of regional economic activity. The last one is the media coverage a location receives, proxied by the number of articles in our dataset that mention the region.

The WBES measures business obstacles by asking firms to evaluate how severely different issues affect their operations. Responses are recorded on a five-point scale ranging from 1, "no obstacle," to 5, "very severe obstacle." [8] Following previous literature, responses such as "does not apply" and "don't know" are coded as 0 (Bhupatiraju, 2020). Table 1 reports the business obstacles considered in the analysis and their corresponding WBES codes.

A key feature of these variables is that they measure firms' perceived operational constraints, not the objective level of each obstacle and not its direct influence on foreign investment. Managers and owners are asked to evaluate the extent to which each obstacle affects their firm's day-to-day operations. These measures therefore capture the perceived local investment climate from the point of view of firms operating in each location.

Including all 15 business obstacles simultaneously in the same regression may create multicollinearity and over-specification problems, because several obstacles are conceptually and empirically related. Previous studies have addressed this issue either by focusing on selected obstacles or by using dimensionality-reduction techniques such as principal component analysis (Kinda, 2010; Bhupatiraju, 2020). We adopt a two-step procedure. First, we use hierarchical clustering to identify groups of correlated obstacles. Second, within each cluster, we apply principal component analysis to construct synthetic investment climate measures.

The hierarchical clustering is based on pairwise correlation-based dissimilarity and Ward's linkage method. This procedure groups obstacles that display similar empirical patterns across firms. The resulting dendrogram, reported in Fig. 2, identifies four clusters. These clusters are not only statistically related but also conceptually interpretable.

The four clusters are as follows. The first cluster, Governance, includes political instability, corruption, tax rates, tax administration, courts, and business licensing and permits. These obstacles capture the institutional, regulatory, fiscal, and political environment in which firms operate. The second cluster, Infrastructure, includes electricity, transport, and customs and trade regulations. These variables reflect the physical and logistical conditions required for business activity. The third cluster, Human capital, includes labor regulations and inadequately educated workforce. These obstacles capture constraints related to labor-market regulation and workforce quality. The fourth cluster, Market frictions, includes practices of competitors in the informal sector, access to finance, access to land, and crime, theft, and disorder. These variables capture competitive, financial, property-related, and security constraints faced by firms.

[8] The exact framing used in the WBES questionnaire is: "Using the response options on the card; To what degree is Transport / Access to Finance / etc… an obstacle to the current operations of this establishment?"

**Table 1**
Business obstacles considered in the analysis and corresponding codes from the World Bank Enterprise Surveys.

| Obstacle | Code in WBES |
|---|---|
| Courts | h30 |
| Tax rates | j30a |
| Tax administration | j30b |
| Business licensing and permits | j30c |
| Political instability | j30e |
| Corruption | j30f |
| Labor regulations | l30a |
| Inadequately educated workforce | l30b |
| Electricity | c30a |
| Transport | d30a |
| Customs and trade regulations | d30b |
| Access to finance | k30 |
| Access to land | g30a |
| Crime, theft and disorder | i30 |
| Practices of competitors in the informal sector | e30 |

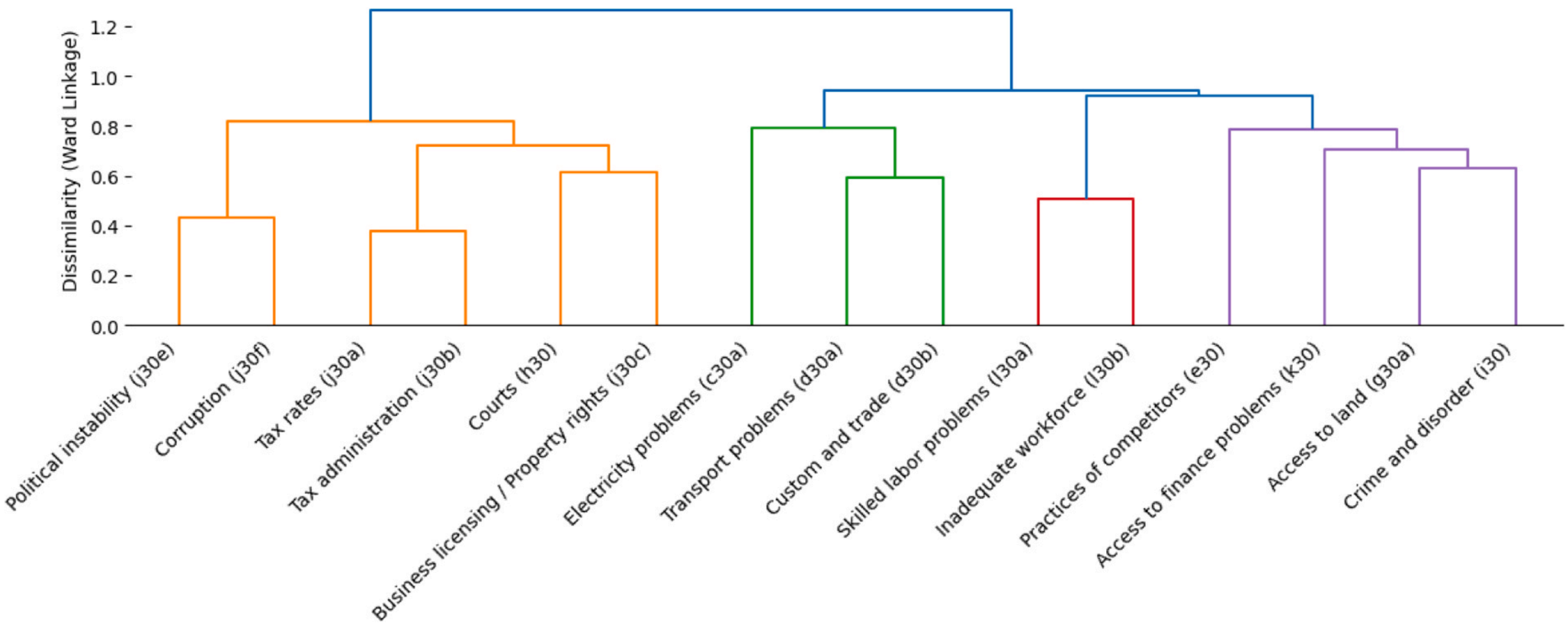

**Fig. 2.** Hierarchical clustering of business obstacles based on correlation dissimilarity and Ward linkage.

After identifying the four clusters, we conduct a principal component analysis separately within each group. We retain components with eigenvalues greater than one, following the Kaiser criterion. In each cluster, the procedure yields a single principal component, suggesting that the obstacles within each group share a common underlying dimension. We then use the component scores as regressors in the empirical analysis. All constituent variables load positively on their respective first principal component; therefore, higher values of each index indicate more severe perceived business constraints in that dimension. The component loadings and proportion of variance explained are reported in Appendix Table A11.

### *3.2. News-based data*

The second data source consists of online news articles collected from Event Registry. Event Registry is a real-time news aggregation and event detection system that collects and processes articles from approximately 75,000 media sources worldwide (Leban et al., 2014). These sources include international and country-specific outlets with broad linguistic and geographic coverage. Event Registry does not produce original content; it aggregates articles from news websites and platforms. Its News Feed service captures between 100,000 and 200,000 articles per day, with English-language articles representing approximately half of the total content (Leban et al., 2014). Event Registry is well-suited to our research design because it provides access to the full text of news articles. This is essential for our empirical strategy, which requires measuring not only whether a theme is mentioned, but also how semantically central that theme is within the article corpus.

We collected 167,746 English-language articles related to industrial investment in Africa, published between 2014 and 2019. We focus on the pre-pandemic period to avoid the influence of COVID-19 on media coverage, investment dynamics, and business conditions, which falls outside the scope of this study. To ensure that the corpus reflects narratives likely to be visible to international investors, we restrict the sample to articles published on high-traffic and reputable websites. Specifically, we select major information websites ranked in the top traffic quintile according to Alexa Internet (Vaughan and Yang, 2013).

We decided to focus on English-language news articles for two reasons. First, previous research has shown that media in the investor's country of origin play an important role in determining cross-country capital flows and investment decisions (Adams, Neumann and Tabrizy, 2018; Fraiberger et al., 2018). We recognize, however, that a large share of investment in Africa may also come from countries where English is not the native language. However, extending the corpus to other languages would have raised several issues, including reduced comparability with previous research that uses English-language news and, from an empirical perspective, the need to adapt the SBS metric to corpora in different languages. Second, the WBES data do not identify the investor's country of origin, information that would have provided a stronger motivation for an analysis differentiated by news language. Future research using alternative datasets could link the share of foreign ownership to news data from the investor's country of origin, potentially exploring heterogeneous effects across languages.

We define relevant articles as those referring simultaneously to industrial activity and investment dynamics. This choice reflects the objective of the study: we do not seek to measure how the media discuss corruption, crime, taxation, infrastructure, or labor issues in general. Instead, we focus on how these themes appear in media coverage relevant to firms, industrial activity, and investment. A broader assessment of the general media representation of these themes across African locations would be valuable, but it lies beyond the scope of this paper.

To identify relevant articles, we designed a Boolean query based on two conceptual dimensions: industrial activity and investment-related dynamics. The industrial activity dimension includes terms such as "production," "fabrication," "factory," "industry," and "manufacture," together with their synonyms and word variations. The investment-related dimension includes terms such as "invest," "FDI," and "development," along with their synonyms and morphological variants. Articles were retained only if they contained at least one term from each conceptual group. This ensures that the selected articles are relevant to both industrial activity and investment.

The keyword list was developed through a literature review and refined through four rounds of validation with three domain experts. Because Event Registry parses the full text of each article, the final corpus was relatively robust to the inclusion or exclusion of individual related terms. Adding additional synonyms did not substantially alter the retrieved set of articles, and omitting single keywords had a limited effect. This robustness reflects the fact that relevant concepts could be detected throughout the full article text rather than only in titles, abstracts, or predefined metadata fields.

### 3.3. Geographic matching between news articles and WBES regions

To link news articles to firm locations, we match article content to the WBES regions identified by the a2x code. This step is necessary because the media-derived variables must be measured at the same geographic level as the locations where firms operate. The WBES provides detailed information on the areas in which firms are surveyed. These areas may correspond to cities, capitals, districts, or combinations of multiple subnational locations, such as Mopti, Ségou, and Sikasso in Mali or Windhoek and Okahandja in Namibia.

Subnational matching is important because media narratives may vary substantially within countries. Issues such as corruption, crime, infrastructure, access to land, and the functioning of the courts may be discussed differently across regions within the same country (Asiedu and Freeman, 2009). However, obtaining reliable subnational measures of these conditions is challenging, particularly in developing countries (Fisman and Svensson, 2007). Linking media content to WBES regions allows us to construct location-specific measures of public narratives around investment-relevant themes.

The geographic matching process faces two main challenges. First, a WBES region may contain several cities, districts, or localities. Matching news articles only to the exact name of the WBES region would therefore exclude articles referring to relevant sub-locations within that region. Second, the same location may be referred to using different spelling conventions, transliterations, or historical names. For example, Fès in Morocco may appear as Fez, Fes, Fassi, Fèz, or Fès. These variations make direct matching difficult.

To address these issues, we constructed a geographic dictionary that links each WBES a2x region to the names of relevant regions, capitals, cities, and other localities contained within it. The dictionary was built using information from Wikidata. We then applied a case-sensitive string-matching procedure to identify mentions of these geographic entities in the title or body of each article. A match required an explicit mention in the title or main text of at least one subnational geographic entity included in the dictionary for that WBES region, while country names alone were not sufficient to assign an article to a region. Each article was linked to one or more WBES regions when at least one geographic entity associated with those regions appeared in the article text. This procedure produces a region-level corpus of news articles that can be combined with firm-level WBES data.

To assess the reliability of the geographic matching procedure, we randomly selected 200 article-region assignments generated by the matching rule and manually verified each assignment. An assignment was classified as correct only when the article explicitly referred to the matched subnational location; mentioning the country alone was not considered sufficient. Of the 200 assignments examined, 191 were correctly matched and nine were false positives, corresponding to a match precision of 95.5 %. Among the 191 verified matches, seven contained the matched location in the article title, while the remaining 184 were identified through the main text. We find that the false positives mainly arise from geographic name ambiguity or lexical overlap. Despite that, given the low observed false-positive rate, we retain the matching procedure without further adjustment, as the remaining classification error is unlikely to significantly affect the results.

Using this procedure, we find that news coverage is not normally distributed across regions. In the matched news corpus, the mean number of articles per region is 18.16 with a standard deviation of 45. The three most-covered regions account for around 27% of all the articles. To alleviate concerns that our results may be driven by regions with high media coverage, we also run a robustness check, described in the results section, that repeats the main analyses by excluding the five regions with the highest media coverage.

### 3.4. Measuring the semantic importance of media themes

To measure the importance of business-related themes in regional media discourse, we use the Semantic Brand Score (SBS), introduced by Fronzetti Colladon (2018). The SBS combines text mining and social network analysis to measure the semantic importance of a concept or theme within a corpus of documents. The method has been applied in several contexts, including the analysis of news articles and financial markets (Fronzetti Colladon et al., 2023).

In this study, the SBS is used to quantify the semantic importance that online news attributes to business-related themes associated with the WBES obstacles reported in Table 1. We start from the WBES obstacle categories because they represent standardized and widely recognized business constraints faced by firms. However, the number of media themes is reduced from 15 to 13. Tax rates and tax administration are grouped under a single taxation theme due to their conceptual similarity. Practices of competitors in the informal sector are excluded because this obstacle is difficult to capture reliably from textual data. The resulting media themes correspond to 13 business-relevant issues.

It is important to clarify that the media variables do not measure the objective severity of the corresponding business obstacles. Instead, they measure the semantic importance of broader themes related to those obstacles in media discourse. For example, the media variable for taxation does not measure the actual tax burden firms face, nor does it capture the exact WBES formulation of tax rates or tax administration. Rather, it measures how central taxation-related language is within media coverage associated with a given region. Another important distinction is that the media variables do not measure investors' actual readership of those frames. This is because we do not have data to measure which news investors read. The choice of the most widely read English-language news outlets can increase the likelihood that investors are exposed to this information, but it cannot provide conclusive evidence of actual readership. In light of this, any interpretation of the media variables should be understood in terms of greater exposure that a potential investor may have to these themes, in line with the previous literature studying the effect of media on investment. These distinctions are important because our theoretical argument concerns the public framing of investment locations, not the direct measurement of local conditions through news articles.

For each theme, we define a set of keywords that represents the underlying concept. The initial keyword lists were derived from the wording of the WBES questionnaire and expanded manually using relevant synonyms and related terms. We then used the SBS BI application (Fronzetti Colladon and Grippa, 2020) to extract additional keywords from the news corpus based on TF-IDF logic (Roelleke and Wang, 2008). Three experts validated the final keyword lists. Table A2 in the Appendix reports the themes analyzed in the study and the main keywords used to compute their SBS.

A central challenge in keyword-based semantic analysis is the risk of false positives. A word may appear in a text without referring to the intended concept. For example, the keyword "land" is relevant to the theme of access to land when it appears in discussions of land availability, ownership, permits, or land-use constraints. However, the same word may appear in unrelated contexts, such as archaeological discoveries or descriptions of natural landscapes. Including all occurrences of "land" would therefore introduce measurement error.

To reduce false positives, we used language models to filter passages unrelated to the theme of interest before computing the SBS. In the case of access to land, for example, the term "land" contributes to the SBS only when it appears in a context related to land access, ownership, availability, or administrative constraints. The same filtering logic is applied to the other themes.[9] This procedure allows us to construct a context-aware measure of semantic importance that goes beyond simple keyword frequency. To validate our methodology, we conduct a manual review on a subset of articles in our dataset. Specifically, we asked a human rater to read each article and indicate whether a given theme was central to the article's narrative. This was done for 130 articles for each of the 13 themes. Annotation results were compared with a binary variable that takes the value 1 if the theme's SBS exceeded the 90th percentile (we used a high threshold because the distributions were skewed). We then calculated the Accuracy and Cohen's kappa to verify the robustness of the results. The average accuracy was 92.07%, while the average Cohen's kappa was 0.5911.

The SBS measures semantic importance along three dimensions: prevalence, diversity, and connectivity. This section provides a conceptual overview of the measure, while the complete formulas are reported in the Appendix. Prevalence captures how frequently a theme appears in the text corpus. Diversity and connectivity require transforming the corpus into a semantic network in which nodes represent words and edges represent co-occurrences between words. Edges are weighted by co-occurrence frequency.

Diversity is measured using distinctiveness centrality, which captures how many different and distinctive words are associated with the theme (Fronzetti Colladon and Naldi, 2020). Connectivity is measured using weighted betweenness centrality, which captures the extent to which the theme acts as a bridge between other words or discourse areas in the semantic network (Brandes, 2001a). The SBS is computed as the sum of the normalized values of prevalence, diversity, and connectivity (Fronzetti Colladon, 2018). A theme, therefore, receives a high SBS when it is frequently mentioned, connected to a heterogeneous and distinctive set of words, and central in linking different parts of the discourse.

[9] To identify these relevant passages within each article, we utilized a straightforward query search algorithm that measures the similarity between a given query and each passage in the text. This process, known as semantic search in text mining literature, aims to retrieve the most pertinent passages based on the query search, similar to how search engines like Google operate. We use a different query search for each theme analyzed. Each query was defined by the list of keywords that describe each topic. To execute this retrieval process, both the query search and news passages are transformed into vectors. We employed the cross-encoder/ms-marco-MiniLM-L6-v2 model, which is widely used for semantic similarity tasks and was selected for its state-of-the-art performance on relevance benchmarks, including NDCG@10 on TREC DL 2019 and MRR@10 on the MS MARCO development set. By representing text as vectors, we can measure the similarity between a query and a passage.

In addition to the SBS, we compute a sentiment measure associated with each theme. Sentiment is calculated using the Loughran and McDonald dictionary, which classifies words as positive or negative in financial and business-related contexts (Loughran and Mcdonald, 2011). For each theme, we identify co-occurring sentiment words and compute the difference between the number of positive and negative words divided by the total number of co-occurring words. This produces a normalized sentiment score that captures the tone of the discourse surrounding each theme. Lastly, from the media content analysis, we also derive the variable N Docs, representing the number of docs at the region-year level.

### 3.5. Temporal alignment between news and firm observations

In the baseline regressions, each observation in our dataset is associated with media variables calculated using the corpus of articles published in the same year. For example, if a firm was interviewed in 2015, all media variables are calculated using news articles published in that same year. The use of contemporaneous media measures is supported by empirical evidence showing that foreign investment can react, within the same period, to changes in the information environment. In the context of political elections, for example, Julio and Yook (2016) show that foreign investment decreases precisely during the election period rather than in the periods preceding it. Jahn and Stricker (2022), distinguishing between pre-election, contemporaneous election, and post-election periods, show that significant adjustments in FDI components occur within the same election quarter.

To determine whether the media effect is lagged and to alleviate concerns regarding reverse causality, we also perform a series of robustness checks using lagged media measures from year $t-1$. In this setting, firms without an available prior-year media measure are omitted from the lagged specification. Results are reported in Table A10.

### 3.6. Econometric Analysis

The empirical analysis proceeds in two steps. First, we examine whether media-derived variables are associated with foreign ownership. Second, we test whether media-derived variables moderate the relationship between perceived business obstacles and foreign ownership.

To estimate the direct association between media framing and foreign ownership, we use the following model:

$$ForeignOwnership_i = \beta_1 SBS_{jcm} + \beta_2 Controls + \gamma_{ck} + \varepsilon \quad (1)$$

where $ForeignOwnership_i$ is the percentage of foreign ownership of firm $i$, located in region $j$, country $c$, and sector $k$. $SBS_{jcm}$ is the Semantic Brand Score of media theme $m$ in region $j$ of country $c$. The vector of controls includes firm age, firm size dummies, agglomeration, region size, regional media coverage, the four investment climate components derived from PCA, and the sentiment associated with the corresponding media theme. We also control for regional economic activity using the median value of night light activity in a region, and for the number of articles mentioning a country (excluding the region of interest) as a proxy for media supply. We include country-sector fixed effects, $\gamma_{ck}$, to absorb unobserved heterogeneity related to national institutions, sector-specific policies, and country-sector economic conditions. Standard errors are clustered at the regional level.

The second model tests whether media frames moderate the association between investment climate and foreign ownership. We estimate:

$$ForeignOwnership_i = \beta_1 X + \beta_2 SBS_{jcm} + \beta_3 (X \times SBS_{jcm}) + \beta_4 Controls + \gamma_{ck} + \varepsilon \quad (2)$$

where $X$ denotes one of the four investment climate components derived from the PCA: governance, infrastructure, human capital, or market frictions. The coefficient of interest is $\beta_3$, which captures whether the association between investment climate constraints and foreign ownership varies with the semantic importance of media theme $m$. This specification allows us to assess whether media narratives amplify or attenuate the relationship between local business obstacles and foreign ownership.

The interpretation of the interaction term depends on the sign and magnitude of $\beta_3$. A negative interaction suggests that the media frame strengthens the negative association between perceived business constraints and foreign ownership. A positive interaction suggests that the media frame weakens, offsets, or reverses that association. This is consistent with our theoretical argument that media frames may either amplify perceived risks or contextualize local constraints in ways that reduce uncertainty. Because the moderation analysis combines 13 media themes with four investment-climate components, it entails 52 interaction tests. To account for multiple testing, we apply the Benjamini-Hochberg false discovery rate procedure jointly to the p-values of all 52 interaction coefficients.

A methodological concern is that perceived business obstacles may be endogenous to foreign ownership. Firms with different productivity, resources, or managerial capabilities may perceive the same local environment differently. For example, a highly productive or well-connected firm may report corruption or infrastructure constraints as minor obstacles, while a less productive firm operating in the same environment may perceive them as severe constraints. If these firm-level perceptions are correlated with unobserved determinants of foreign ownership, ordinary least squares estimates may be biased.

To address this concern, we follow previous research using instrumental variables for firm-reported business constraints (Honorati and Mengistae, 2010; Kinda, 2010; Aterido, Hallward-Driemeier and Pagés, 2011; Bhupatiraju, 2020). Specifically, we instrument each firm-level investment climate component using the average value of the same component among other firms operating in the same region-sector cell. To avoid mechanical correlation between a firm's own response and the instrument, the average is constructed using a leave-one-out procedure. The intuition is that the average perception of similar firms operating in the same local and sectoral environment captures common local constraints, while excluding the idiosyncratic perception of the focal firm. We estimate the models using two-stage least squares.

### 3.7. Dealing with Media Variables Endogeneity

A potential concern with the baseline estimates is that the SBS variables may be endogenous to foreign ownership. Media coverage may respond to the presence of foreign investors, or both media representation and foreign ownership may be jointly determined by unobserved characteristics of the locations in which firms operate. To address this concern, we construct a set of Bartik-style instruments that combine a common, plausibly exogenous shock to media attention with cross-country differences in historical exposure to each media frame. This approach is related to the broader shift-share literature, in which heterogeneous exposure shares are interacted with common shocks (Goldsmith et al., 2020; Borusyak et al., 2022). In particular, Borusyak et al. (2022) show that identification in shift-share designs can be based on the quasi-random assignment of the underlying shocks, as we explain below.

In our setting, the shift exploits the occurrence of major international sporting events: the Olympic Games and the FIFA World Cup. Over our sample period, these events occurred in 2014, 2016, and 2018, corresponding to the 2014 and 2018 FIFA World Cups and the 2016 Summer Olympic Games. Ideally, the shift should affect our media variable without directly influencing foreign ownership. This identifying mechanism builds on the idea of news crowding out. Media outlets face a finite amount of space and audience attention. Large news events absorb a substantial share of this attention and consequently reduce the coverage devoted to competing topics. Eisensee and Strömberg (2007), for example, exploit the Olympic Games as a source of exogenous variation in news pressure and show that natural disasters receive less television coverage when they coincide with the Games. More generally, Durante and Zhuravskaya (2018) exploit predictable high-profile events to generate exogenous variation in the availability of news space. We apply the same intuition to our setting: during years marked by major sporting events, less media attention is available to other topics.

To obtain cross-sectional variation in exposure to this common shock, we construct a predetermined measure of each country's historical propensity to receive coverage related to a given theme. For each country in our dataset, we collect information about the number of articles published in English by major global news outlets for each topic analyzed between 1980 and 2013, using LexisNexis. The number of articles for each topic is obtained by entering the same set of keywords used to calculate the SBS values into the LexisNexis search query. We standardized each measure by the total number of articles for a given country.

This metric captures the historical importance of theme $(m)$ in the international media coverage of country $(c)$, while normalizing for persistent differences in the overall amount of coverage received by different countries. LexisNexis is particularly suitable for this purpose because Event Registry started collecting data in 2014, whereas constructing the exposure measure requires a substantially longer pre-sample period. Moreover, this exercise does not require downloading the full text of individual articles: LexisNexis searches the full text of its archived news content, so the number of articles satisfying each country-theme query is sufficient to construct the exposure measure.

The instrument for media theme $m$ in country $c$ and year $t$ is defined as:

$$Z_{cmt} = Sport_t \times Exposure_{cm}$$

where $Sport_t$ equals 1 if in the corresponding year the sports event occurred and $Exposure_{cm}$ is the historical exposure for country $c$ and theme $m$. The intuition is the following. Major sporting events generate a common negative shock to the amount of space available for non-sport news, but the effect of this shock on the SBS of a given theme need not have a predetermined sign. If a theme has historically occupied an important position in a country's coverage, the sporting-event shock may reduce coverage of that theme, lowering its SBS. However, it may also reduce competing news more strongly, leaving the remaining coverage relatively more concentrated around the historically salient theme and therefore increasing its SBS. In either case, countries with greater historical exposure to a theme should experience a larger change in its semantic importance. The interaction, therefore, converts a common time shock into heterogeneous cross-country variation in media framing, without imposing an a priori direction on this response.

Formally, for firm $i$, located in region $j$, country $c$, and observed in year $t$, the first-stage equation is

$$SBS_{jcmt} = \pi Z_{cmt} + \rho Exposure_{cm} + \varphi Controls + \delta_{kt} + \varepsilon$$

where $SBS_{jcmt}$ denotes the observed SBS, $Z_{cmt}$ is the Bartik-style instrument, $Exposure_{cm}$ is the predetermined historical exposure measure and $\delta_{kt}$ denotes industry-by-year fixed effects. The fixed effects absorb shocks that are common to firms operating in the same industry in a given year, including the aggregate time component of the sporting-event shock.

An important feature of our data is that each country is observed in only one survey year. Although the dependent variable is measured at the firm level and the media variables vary across subnational locations, the instrument varies at the country level within each survey year. The data, therefore, do not provide within-country variation over time. Country fixed effects cannot be included in this specification because they would absorb the cross-country variation that identifies both historical exposure and its interaction with the sporting-event shock.

We account for observable cross-country heterogeneity through a set of country-level macroeconomic controls. Specifically, we control for annual CPI inflation, total natural resource rents as a share of GDP, net FDI inflows as a share of GDP, GDP in current US dollars, and domestic credit to the private sector as a share of GDP. Inflation, measured as the annual percentage change in the consumer price index, captures differences in macroeconomic stability. Natural resource rents measure the combined rents generated by oil, natural gas, coal, minerals, and forests relative to GDP and account for differences in countries' dependence on resource-based economic activity, which may, in turn, affect foreign investment. Net FDI inflows as a share of GDP capture broader contemporaneous cross-country differences in aggregate inward foreign investment. GDP in current US dollars controls for differences in market size, while domestic credit to the private sector, measured as the financial resources provided to the private sector by financial corporations relative to GDP, captures differences in domestic financial development. All five variables are obtained from World Bank data. The inflation series is based on IMF International Financial Statistics, the natural-resource-rents measure draws on the World Bank's Changing Wealth of Nations estimates, and the FDI series primarily relies on IMF International Financial Statistics and Balance of Payments data. GDP is based on national accounts data from national statistical authorities, central banks, and the OECD, supplemented where necessary by World Bank staff estimates. Domestic credit to the private sector is primarily based on IMF International Financial Statistics, with GDP data from the World Bank and OECD used to construct the ratio.

The sign of $\pi$ is theoretically ambiguous. A negative coefficient would indicate that the sporting-event shock reduces the SBS more strongly in countries with greater historical exposure to that frame. A positive coefficient may instead arise if the reduction in overall non-sport coverage disproportionately affects competing topics, making a historically salient frame relatively more prominent among the articles that remain. The relevant first-stage prediction is therefore not $\pi < 0$ but $\pi \neq 0$. Identification comes from the differential response to the common sporting-event shock across countries with different predetermined levels of historical exposure.

The second stage relates the instrumented media variable to firm-level foreign ownership:

$$ForeignOwnership_i = \beta\widehat{SBS} + \alpha Exposure_{cm} + \theta Controls + \delta_{kt} + \varepsilon$$

**Table 2**

Regression results of foreign ownership and media-derived variables. Standard errors in parentheses. Significance levels: p < 0.1 (*), p < 0.05 (**), p < 0.01 (***).

| | Dependent variable: % of Foreign Ownership | | | |
|---|---|---|---|---|
| | (1) | (2) | (3) | (4) |
| SBS Access to land (g30a) | 0.184** | | | 0.125** |
| | (0.080) | | | (0.050) |
| SBS Crime, theft and disorder (i30) | | -4.331*** | | -3.864*** |
| | | (1.029) | | (0.943) |
| SBS Inadequately educated workforce (l30b) | | | 0.192*** | 0.119* |
| | | | (0.060) | (0.064) |
| Age (ln) | -0.008 | -0.008 | -0.008 | -0.008 |
| | (0.010) | (0.010) | (0.010) | (0.010) |
| Sales Growth | 0.000 | 0.000 | 0.000 | 0.000 |
| | (0.000) | (0.000) | (0.000) | (0.000) |
| Employee Growth | -0.000 | -0.000 | -0.000 | -0.000 |
| | (0.001) | (0.001) | (0.001) | (0.001) |
| Dummy Small | -0.087*** | -0.086*** | -0.086*** | -0.086*** |
| | (0.020) | (0.020) | (0.020) | (0.020) |
| Dummy Big | 0.127*** | 0.126*** | 0.126*** | 0.125*** |
| | (0.034) | (0.034) | (0.034) | (0.034) |
| Market Frictions | -0.018** | -0.018** | -0.019** | -0.019** |
| | (0.008) | (0.008) | (0.008) | (0.008) |
| Infrastructure | 0.021** | 0.021** | 0.021** | 0.021** |
| | (0.011) | (0.011) | (0.011) | (0.011) |
| Human Capital | -0.007 | -0.007 | -0.007 | -0.007 |
| | (0.009) | (0.009) | (0.009) | (0.009) |
| Governance | 0.001 | 0.001 | 0.001 | 0.001 |
| | (0.007) | (0.007) | (0.007) | (0.007) |
| Agglomeration | 0.000*** | 0.000*** | 0.000*** | 0.000*** |
| | (0.000) | (0.000) | (0.000) | (0.000) |
| Region Size | -0.001*** | -0.001** | -0.001*** | -0.000** |
| | (0.000) | (0.000) | (0.000) | (0.000) |
| N Docs | 0.001*** | 0.001*** | 0.001*** | 0.001*** |
| | (0.000) | (0.000) | (0.000) | (0.000) |
| Nightlight Activity | 0.004** | 0.004** | 0.003** | 0.003** |
| | (0.002) | (0.002) | (0.001) | (0.001) |
| Country Coverage | 0.069 | 0.533 | 0.056 | 0.032 |
| | (0.332) | (0.323) | (0.345) | (0.355) |
| Sentiment | 0.258* | 0.168 | 0.363** | 0.224** |
| | (0.149) | (0.154) | (0.141) | (0.094) |
| Constant | 0.197*** | 0.191*** | 0.196*** | 0.193*** |
| | (0.031) | (0.031) | (0.031) | (0.031) |
| Observations | 6232 | 6232 | 6232 | 6232 |
| R Squared | 0.38 | 0.38 | 0.38 | 0.38 |
| N Countries | 29 | 29 | 29 | 29 |
| N Regions | 107 | 107 | 107 | 107 |

where $ForeignOwnership_i$ denotes the share of foreign ownership in firm $i$, and $\widehat{SBS}$ is the predicted SBS measure obtained from the first stage. The historical exposure measure and the same set of controls and industry-by-year fixed effects included in the first stage are also retained in the second stage.

The identifying assumption is not that historical differences in media coverage across countries are randomly assigned. Historical exposure may reflect persistent economic, institutional, or other country characteristics. Rather, identification relies on the timing of the sporting-event shock being unrelated to contemporaneous country-specific determinants of foreign ownership and, conditional on historical exposure, industry-by-year fixed effects, and the observed controls, affecting foreign ownership only through its differential impact on media attention. Constructing $Exposure_{cm}$ exclusively from the 1980–2013 period further ensures that the exposure component cannot mechanically respond to foreign ownership or media outcomes observed during the estimation period.

Given the limited amount of country-level identifying variation and the resulting concern about weak instruments, we complement conventional IV inference with Anderson-Rubin inference, which remains valid under weak identification.

## 4. Results

This section presents the empirical results in two steps. First, we examine whether the SBS of different media themes is associated with foreign ownership. Second, we investigate whether the SBS moderates the relationship between firms' perceived business obstacles and foreign ownership. Because the SBS measures of different themes may be correlated, for example, media discussions of corruption may co-occur with crime, courts, or political instability, we estimate the models separately for each SBS theme. We then estimate specifications including the relevant SBS measures jointly to verify that the results are not driven by multicollinearity among media variables.

### 4.1. Media frames and foreign ownership

Table 2 reports the results of the first set of regressions, where the dependent variable is the percentage of foreign ownership. Among the thirteen media themes analyzed, three display a statistically significant association with foreign ownership: access to land, crime, theft and disorder, and inadequately educated workforce. Columns 1–3 report the results for each SBS separately. Column 4 includes the three SBS jointly. The maximum Variance Inflation Factor in the joint specification is 3.78, suggesting that multicollinearity is not a major concern. The coefficients also remain relatively stable across specifications.[10]

The results show that the semantic importance of land-related and workforce-related themes is positively associated with foreign ownership. In Column 1, the coefficient of SBS access to land is positive and statistically significant. This association remains positive and significant in the joint specification reported in Column 4. Similarly, the SBS of inadequately educated workforce is positively associated with foreign ownership in Column 3 and remains statistically significant in Column 4. By contrast, the SBS of crime, theft and disorder is negatively associated with foreign ownership.[11] However, as shown in the robustness checks, the direction of the social safety coefficient is sensitive to the temporal ordering and identification strategy (see Table 3 and Table A10).

To verify whether the observed media frame effect is driven primarily by regions with high media coverage, we repeat the analysis by removing the five regions with the highest media coverage. The direction of the results remains consistent with the main findings. The only difference we report is a loss of significance for the workforce theme in the joint specification. One reason could be a loss of statistical power resulting from the smaller number of observations.[12]

A one standard deviation increase in access to land is associated with an increase of 1.3 %age points in foreign ownership, and almost 2 %age points when the theme is related to the workforce. On the other hand, a one SD increase in the social safety theme is associated with a decline in foreign ownership of 2.6 %age points. These values are approximately 13–25% of the mean of foreign ownership (equal to 0.104).

To test whether the media channel operates through the intensive or extensive margin, we present two regression analyses in the appendix using alternative specifications. In the first specification, we use only observations with foreign ownership greater than 0, with the percentage of foreign ownership as the dependent variable, while in the second specification, we use the complete set of observations and estimate a linear probability model with a dummy variable as the dependent variable that takes the value 1 if the percentage of foreign ownership exceeds 10%, in line with previous literature (Kinda, 2010) (see Table A7). The results show that the analyzed frames operate on two distinct margins. The access to land coefficient is significant on the intensive margin but not on the extensive margin. This result does not change when we use a logit model. On the other hand, the other two themes are significant on the extensive margin but not on the intensive one. This points out that, in addition to having a heterogeneous association with foreign capital, media frames are associated with foreign ownership on different margins, depending on the theme being covered.

The results described so far also show that the SBS coefficients remain statistically significant after controlling for sentiment. This suggests that the association between media frames and foreign ownership is not driven by whether coverage is positive or negative.

[10] When we include all thirteen SBS themes in a single regression, the maximum VIF increases substantially to 10.19, indicating severe multicollinearity. For this reason, we do not report that specification.

[11] Since the dependent variable is bounded between 0 and 1, we repeat the analysis using a fractional-reg logit model as a robustness. The sign and magnitude of the coefficients are consistent with those of the main results.

[12] We repeat the analysis with outlier removal and obtain consistent results.

**Table 3**
Instrumental variable regression results.

| | (1) | (2) | (3) |
|---|---|---|---|
| | Access to land | Crime, theft and disorder | Inadequately educated workforce |
| SBS Access to land | 2.419*** | | |
| | (1.156) | | |
| SBS Crime, theft and disorder | | 25.275*** | |
| | | (12.044) | |
| SBS Inadequately educated workforce | | | 0.741 |
| | | | (0.633) |
| Observations | 5632 | 5632 | 5632 |
| First-stage coefficient | -0.249 | 0.015 | -0.213 |
| First-stage s.e. | 0.079 | 0.005 | 0.060 |
| First-stage p-value | 0.002 | 0.002 | 0.000 |
| K-P rk Wald F | 10.06 | 9.39 | 12.50 |
| K-P rk LM p-value | 0.015 | 0.039 | 0.020 |
| Anderson-Rubin p-value | 0.009 | 0.002 | 0.157 |
| Controls | Yes | Yes | Yes |
| Industry-year f.e. | Yes | Yes | Yes |
| S.e. clustered by | Country | Country | Country |
| Country clusters | 26 | 26 | 26 |

The control variables behave largely in line with expectations. Larger firms display higher levels of foreign ownership, and agglomeration is positively and consistently associated with foreign ownership, indicating that foreign capital is more likely to be present in region-sector environments where other foreign-owned firms are already located. Region size is negatively associated with the dependent variable. Conditional on other factors, foreign ownership is not simply concentrated in regions with more surveyed firms, while employee growth is not significantly associated with it.

The investment climate components also show some meaningful associations. Market frictions are negatively and significantly associated with foreign ownership across all specifications; in contrast, infrastructure displays a positive association with it. These patterns are broadly consistent with previous work using firm-level measures of business obstacles, which shows that different types of constraints may have heterogeneous associations with the presence of foreign capital. For example, Bhupatiraju (2020) finds that some infrastructure-related obstacles can be positively associated with inward FDI, while access to finance and other market-related frictions may discourage foreign presence.

As additional analyses, we also test alternative measures of media importance and decompose the SBS into its three components, prevalence, diversity, and connectivity. These additional analyses are reported in Appendix Tables A8 and A9.

To reduce concern about omitted variable bias, in Table 3 we report the regression results obtained using the instrumental variable strategy. The first-stage results suggest that the instruments are reasonably relevant, with Kleibergen-Paap F-statistics close to or above the conventional threshold of 10. The underidentification test is also rejected in all three specifications. Given the potential concerns about instrument strength, statistical significance is assessed using Anderson-Rubin p-values, following Keane and Neal (2024), who recommend weak-instrument-robust inference rather than conventional 2SLS *t*-tests. In this specification the number of countries falls from 29 to 26 because three countries do not report data for domestic credit to the private sector as a share of GDP, which is included among the controls.

The coefficients for access to land and crime, theft and disorder are positive and statistically significant, whereas the coefficient for inadequately educated workforce is not statistically significant. Notably, the coefficient for crime, theft and disorder changes sign relative to the contemporaneous baseline specification, becoming positive in the IV estimates. This sign reversal is also observed when using lagged media measures (see Table A10), suggesting that the negative baseline association is not robust to alternative strategies addressing temporal ordering and endogeneity. The relationship between social safety narratives and foreign ownership, therefore, warrants further investigation, although these alternative specifications point toward a positive association.

### *4.2. The moderating role of media frames*

We carry out a second set of regressions to examine whether media frames moderate the relationship between perceived business obstacles and foreign ownership. Given the large number of possible interactions between media themes and investment climate components, Fig. 3 displays the statistically significant interaction coefficients after applying the Benjamini-Hochberg false discovery rate correction jointly to all 52 tests. The complete regression results are reported in Table A6, Panel A, in the Appendix. Each cell in Fig. 3 indicates whether the relationship between a given business obstacle component and foreign ownership is significantly moderated by the SBS of a specific media theme and its direction. Grey cells represent no significant interaction.

We find that only one of the thirteen frames analyzed, access to justice, negatively moderates the relationship between business obstacles and foreign ownership. More specifically, this moderating effect emerges for obstacles related to Infrastructure and Human Capital. As shown in the two panels of Figure A1, when SBS is low, neither Infrastructure nor Human Capital obstacles are significantly negatively associated with foreign ownership. As SBS increases, however, the relationship becomes more negative. Only in regions where SBS exceeds the threshold of approximately 0.08 the relationship between the obstacles and the presence of foreign capital becomes significantly negative. At these points, a one-standard-deviation increase in Infrastructure is associated with a 6.3 %age-point decline in our dependent variable, while a one-standard-deviation increase in Human Capital is associated with a 6.5 %age-point decline in the dependent variable. These values represent approximately 20% of one standard deviation of foreign ownership. These results remain robust when we use the individual components instead of the PCA factors (results with individual components are reported in panel B of Table A6). In contrast, we do not observe statistically significant interactions between this media theme and other obstacles.

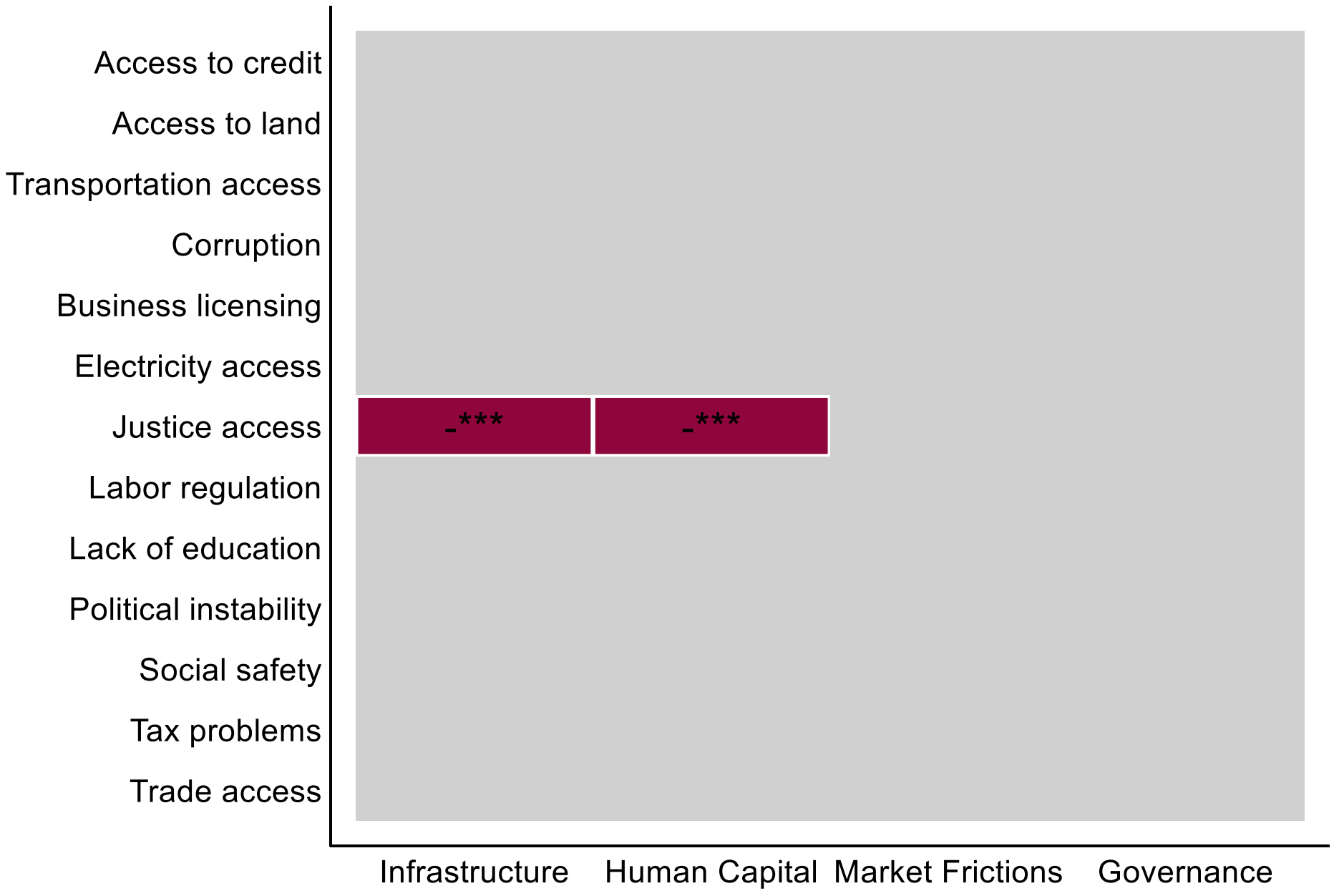

**Fig. 3.** Significant interaction coefficients between media themes and investment climate components. Complete regression results are reported in Table A6 Panel A. + /- = coefficient sign. * $p < 0.10$, ** $p < 0.05$, *** $p < 0.01$ after BH correction.

## 5. Discussion

The first argument that emerges from our results is that improvements in a location's economic and institutional fundamentals remain key prerequisites for foreign investment. When market frictions are present in a location, it becomes difficult to attract foreign investments. On the other hand, agglomerations of foreign companies or areas with higher levels of economic and productive activity can attract a larger volume of foreign capital. Fundamentals at the individual-company level remain central, for example, company size. The larger a firm, the greater the likelihood of foreign capital. These results, therefore, point in one direction: economic and institutional improvements remain essential for attracting investments.

Our second argument is linked to the main focus of our research: the role of the media. We find that the semantic importance of land-related frames is positively associated with foreign ownership. More broadly, when the lagged and IV specifications are also considered, the direct associations between media variables and foreign ownership are generally positive. However, the contemporaneous estimate for social safety is negative. These findings can be interpreted through two complementary mechanisms. First, media coverage may increase the visibility of locations that are relevant to foreign investors. In line with the literature on foreign location choice, locations that are more visible in public information channels are more likely to be considered by investors (Kulchina, 2014). Second, media coverage may reduce information asymmetries by providing publicly available information about local conditions (Chen et al., 2021; Aman and Moriyasu, 2022). This mechanism is particularly relevant where official data are limited or difficult to compare across locations. Even when such information is imperfect, it can reduce uncertainty about local conditions and help investors form expectations about investment opportunities. The scope of this interpretation should, however, reflect the information environment captured by our data. Specifically, our results concern media frames circulating through high-traffic English-language online news and should not be generalized to local-language, smaller-scale, or offline information channels through which investors may also acquire information.

A reduction in information asymmetry may also help explain the results of the interaction analysis. Investment issues become negatively associated with foreign presence when media coverage focuses on legal problems. This may happen because investors have more information about major investment barriers and therefore decide not to invest. This does not occur, however, when the frames are related to other issues, or simply when media coverage of a region intensifies. This is puzzling and invites further research. On the one hand, the strong negative interaction we find between SBS justice and Infrastructure and Human Capital constraints can be explained by a lower degree of information asymmetry. Investors become more aware of a location's problems and obstacles and may decide to invest less. On the flip side, the media can create an image in investors' minds that may act as a disincentive to investment (Entman, 1993). Distinguishing between these two channels is certainly an important avenue for future research.

In light of these results, our last argument stresses the importance of institutional reforms. Our findings suggest that the media matters. Beyond the mere media coverage a location receives, the frame through which it is portrayed can lead to heterogeneous effects and outcomes. This reinforces the concept that institutional reforms remain central and that biases in media reporting, which may not genuinely reflect such reforms, can produce unintended effects on a location's development. Media content can be the result of promotional campaigns, selective reporting, and other biases. Because of this, greater cooperation among investment promotion agencies and all local institutions that provide information, such as local government agencies, independent media, and data providers becomes essential. Such collaborations can lead to more accessible and credible information and, consequently, accurately reflect institutional reforms.

## 6. Implications and conclusions

Foreign investment, particularly in manufacturing, is widely regarded as an important driver of structural transformation, employment creation, technology transfer, and economic development in emerging economies. In Africa, however, industrial development remains constrained by several structural barriers, and foreign manufacturing investment continues to account for only a limited share of total greenfield investment (UNCTAD, 2019a). Understanding the factors associated with foreign ownership across African regions is therefore important for both academic research and policy.

This paper contributes to this debate by examining the association between media frames and foreign ownership at the subnational level in Africa. In financial markets, public news and advertising can draw investors' attention to salient firms, increasing trading activity and, in some cases, short-run returns (Nofsinger, 2001; Fehle et al., 2005). The tone of financial coverage can also exert temporary pressure on stock prices (Tetlock, 2007), while evidence from other settings shows that media exposure can alter individual beliefs and perceptions (DellaVigna and Kaplan, 2007; Rizzica and Tonello, 2015). However, relatively little attention has been paid to how the framing of locations in the media is associated with foreign ownership. Most existing studies focus on media visibility, press freedom, or aggregate measures of information availability. This paper complements these approaches by examining the semantic importance of business-related frames in the media discourse. Using firm-level data from the World Bank Enterprise Surveys and a large corpus of online news articles, the paper shows that media frames are associated with foreign ownership in a theme-specific way. The frame that shows the strongest (positive) correlation with the presence of foreign capital in a location is access to land. Whereas the frame that most strongly moderates (negatively) the relationship between investment climate constraints and foreign ownership is the one related to legal issues.

Our findings suggest that improving a location's economic and institutional conditions remains essential for attracting foreign investment. This means that policy efforts aimed at attracting foreign capital should begin with the context in which such investments take place. Our focus on the information environment shows that, while stronger institutions cannot be replaced by communication, clear and credible media coverage can make it easier for foreign investors to observe and evaluate them. Our estimates do not indicate that such media coverage directly influences investment decisions. We do, however, find a positive correlation between foreign ownership and certain issues highlighted by the media. We also find that the association between business constraints and foreign ownership can become significant depending on the issues highlighted in regional news reports. In this sense, regional reporting may matter insofar as it makes implemented changes in the local business environment visible and understandable to foreign investors, rather than as a substitute for those changes themselves.

Our findings may also be relevant to investment promotion agencies (IPAs) and development organizations operating in Africa. Investment promotion is not limited to advertising a location to foreign firms; it also involves reducing information asymmetries, facilitating interactions between investors and public authorities, and promoting improvements in the investment climate. These functions can be particularly valuable in developing economies, where information and administrative frictions tend to be greater (Harding and Javorcik, 2011; World Bank Group, 2020). In the East African Community, for example, IPAs regularly provide information on legal and administrative procedures. At the same time, one-stop shops are tasked with assisting investors, among other things, in acquiring land and public utilities (OECD, 2024). In this context, our findings do not imply that investment promotion agencies should seek to influence media coverage. Rather, they suggest that information and facilitation services for investors could benefit from being organized around the specific issues most relevant to the relationship between local conditions and foreign ownership. For development organizations, these findings may additionally provide a diagnostic perspective for the design of technical assistance. Media-based measures can help identify the issues through which local investment environments are publicly represented and, in combination with conventional measures of the business climate, indicate areas in which improvements in institutions have not yet been accompanied by equally clear and accessible information. This can support investment portals, digital one-stop shops, regulatory transparency, and the capacity of national and subnational investment promotion agencies. However, because the relationships documented in this paper are associational, media frames should not themselves be interpreted as policy targets or measures of investment-promotion performance. Their main policy value lies instead in identifying informational dimensions that may complement substantive reforms to the local investment environment.

Several limitations should be acknowledged. First, our analysis focuses exclusively on English-language online news published by high-traffic sources and does not directly observe investors' media consumption. This choice is consistent with our focus on foreign investors and allows us to examine a segment of the international information environment that previous research suggests may be particularly relevant for cross-border investment, but our data do not reveal whether individual foreign investors actually read, trust, or respond to these sources. Accordingly, the media variables should be interpreted as measures of the international English-language information environment surrounding a location, rather than as proxies for investor attention or direct media exposure. Local-language media, smaller local outlets, and offline information channels may convey different narratives and may also reach foreign investors indirectly through local partners, advisors, or intermediaries. Our findings should therefore be interpreted as referring specifically to the high-traffic English-language online news environment represented in our corpus, rather than as evidence about the entire media landscape. Future research could extend the analysis to different languages and source origins and examine whether the relationship between media framing and foreign ownership varies across international, national, and local information environments. Future research could also combine media-derived measures with investor-level data on information acquisition or media consumption and extend the analysis to local-language sources.

Second, the article selection strategy focuses on news related to industrial activity and investment. This choice is consistent with the study's purpose, but it excludes broader media discussions that may also influence perceptions of African locations. Future studies could expand the corpus to include a wider range of economic, political, and social topics. Third, although the semantic measures capture the importance of business-related themes in media discourse, they do not measure the objective severity of the corresponding business obstacles. The SBS variables should therefore be interpreted as measures of media framing, not as direct measures of local conditions. Future research could combine media-derived indicators with independent subnational measures of infrastructure, crime, governance, or labor-market conditions. Fourth, the WBES outcome records each firm's foreign equity share as of the survey date. It therefore captures current foreign-capital presence and intensity rather than new FDI flows, entry events, or location choices. Although the WBES reports firm age, it does not identify when foreign ownership began. Firm establishment dates, therefore, cannot provide a valid pre-entry media window because foreign owners may have acquired equity after the firm was established. Combined with the cross-sectional design, this limitation means that the analysis identifies associations and moderation patterns rather than causal effects on investment decisions. Moreover, while we try to address endogeneity in the interaction analysis by instrumenting the business-obstacle component, jointly instrumenting the business obstacle, the media variable, and their interaction would require substantially greater statistical power than is available in our setting; future research could therefore develop alternative identification strategies to more robustly isolate the causal component of these interaction effects. Longitudinal ownership data or project-level entry dates would allow future research to align media exposure with subsequent foreign entry.

Despite these limitations, the paper provides new evidence on the relationship between media narratives and foreign ownership in Africa. It shows that media coverage matters not only as a source of visibility but also as a framing mechanism. By shaping how locations are associated with land access, workforce conditions, crime, courts, taxation, and infrastructure, media narratives contribute to the informational environment in which foreign investors evaluate local risks and opportunities. The findings, therefore, suggest that understanding foreign investment in developing economies requires attention not only to economic fundamentals and institutional constraints, but also to the public narratives through which places are represented.

## CRediT authorship contribution statement

**Andrea Fronzetti Colladon:** Writing – review & editing, Writing – original draft, Software, Methodology, Data curation, Conceptualization. **Roberto Vestrelli:** Writing – review & editing, Writing – original draft, Methodology, Formal analysis, Data curation, Conceptualization. **Stefania Bait:** Writing – original draft, Conceptualization. **Massimiliano Maria Schiraldi:** Writing – review & editing.

## Declaration of Competing interest

None.

## Appendix

Table A1 reports the 200 most frequent URLs of news websites in our dataset.

**Table A1**
200 most frequent news sources

| Sources | | |
|---|---|---|
| vanguardngr.com | thehill.com | uk.finance.yahoo.com |
| ghanaweb.com | ibtimes.co.uk | metro.co.uk |
| punchng.com | cnn.com | euronews.com |
| seekingalpha.com | npr.org | tabnak.ir |
| dailymail.co.uk | au.finance.yahoo.com | newsroom.ucla.edu |
| reuters.com | time.com | jalopnik.com |
| nation.co.ke | foxnews.com | europe.chinadaily.com.cn |
| marketwatch.com | rt.com | articles.economictimes.indiatimes.com |
| forbes.com | in.investing.com | usa.chinadaily.com.cn |
| wsj.com | abcnews.go.com | newsinfo.inquirer.net |
| bloomberg.com | chicagotribune.com | uk.businessinsider.com |
| economictimes.indiatimes.com | news.com.au | news.sky.com |

(continued on next page)

**Table A1** (*continued*)

| Sources | | |
|---|---|---|
| theguardian.com | hindustantimes.com | ph.news.yahoo.com |
| standardmedia.co.ke | newyorker.com | aljazeera.com |
| independent.co.uk | hollywoodreporter.com | news.abs-cbn.com |
| xinhuanet.com | cbsnews.com | msnbc.com |
| nytimes.com | nbcnews.com | investopedia.com |
| ft.com | inc.com | cidrap.umn.edu |
| washingtonpost.com | state.gov | africa.chinadaily.com.cn |
| huffingtonpost.com | wired.com | opinion.inquirer.net |
| fool.com | europarl.europa.eu | sports.yahoo.com |
| msn.com | yenisafak.com | cbssports.com |
| nairaland.com | asia.nikkei.com | news.sap.com |
| cnbc.com | en.farsnews.com | tasnimnews.com |
| politico.com | denverpost.com | gizmodo.com |
| timesofindia.indiatimes.com | breitbart.com | e.vnexpress.net |
| moneycontrol.com | the-star.co.ke | ec.europa.eu |
| telegraph.co.uk | entrepreneur.com | voices.nationalgeographic.com |
| news.yahoo.com | pbs.org | sg.news.yahoo.com |
| finance.yahoo.com | profit.ndtv.com | movieweb.com |
| af.reuters.com | english.alarabiya.net | washpost.bloomberg.com |
| usatoday.com | ndtv.com | allure.vanguardngr.com |
| chinadaily.com.cn | cnet.com | admin.punchng.com |
| thehindu.com | usnews.com | koreajoongangdaily.joins.com |
| qz.com | news18.com | arstechnica.com |
| firstpost.com | chron.com | news.un.org |
| in.reuters.com | mirror.co.uk | engadget.com |
| businessinsider.com | business.inquirer.net | avclub.com |
| smh.com.au | eeas.europa.eu | gadgets.ndtv.com |
| bbc.com | globaltimes.cn | techradar.com |
| un.org | vox.com | businesstoday.intoday.in |
| yahoo.com | nypost.com | tribune.com.pk |
| sputniknews.com | chinapost.nownews.com | tech.firstpost.com |
| zeenews.india.com | billboard.com | vice.com |
| economist.com | dawn.com | ew.com |
| edition.cnn.com | indiatoday.intoday.in | sciencedaily.com |
| legit.ng | uk.investing.com | bleacherreport.com |
| markets.businessinsider.com | nymag.com | ca.news.yahoo.com |
| dw.com | naij.com | digitaltrends.com |
| indianexpress.com | rappler.com | english.rfi.fr |
| variety.com | thesun.co.uk | english.farsnews.com |
| latimes.com | fortune.com | deadspin.com |
| au.news.yahoo.com | thedailybeast.com | comicbook.com |
| in.news.yahoo.com | motherboard.vice.com | complex.com |
| mobile.nation.co.ke | screenrant.com | pcworld.com |
| uk.reuters.com | spiegel.de | ansa.it |
| investing.com | wnd.com | nationalgeographic.com |
| abc.net.au | oneindia.com | en.dailypakistan.com.pk |
| sfgate.com | news.nationalgeographic.com | westernjournal.com |
| theatlantic.com | news.blogs.nytimes.com | infowars.com |
| nature.com | rollingstone.com | republika.co.id |
| rediff.com | stream.marketwatch.com | slate.com |
| urdupoint.com | espn.com | mlb.com |
| techcrunch.com | mashable.com | indiatimes.com |
| zerohedge.com | dealbook.nytimes.com | goal.com |
| theepochtimes.com | theverge.com | blogs.ft.com |
| cbc.ca | uk.news.yahoo.com | |

Table A2 contains the name of each SBS theme analyzed. For each theme, in column 2, we provide the title of a randomly selected article. To select each title, we pick a random article from the set of articles with higher SBS (top decile) for the respective theme. The third column contains the main keywords that were used to calculate the SBS in the content analysis.

**Table A2**
Title of each theme analyzed in the news corpus, alongside its corresponding WBES code

| SBS Cluster | Sampled News Title | Main Keywords |
|---|---|---|

(*continued on next page*)

**Table A2** (*continued*)

| SBS Cluster | Sampled News Title | Main Keywords |
|---|---|---|
| Courts (h30) | Experts push for better access to legal assistance for vulnerable people in Africa. | judiciary, court, justice system, legal process, judgment, lawsuit, rule of law, magistrate, litigation, high court, supreme court, tribunal, legal rights, case ruling, prosecution, legal framework, judicial reform |
| Taxes (j30a and j30b) | Experts criticise 'highest tax rate in Africa', call for urgent. | tax, taxation, revenue, fiscal policy, tax evasion, tax base, taxpayer, customs duty, VAT, excise, corporate tax, personal income tax, tax reform, tax compliance, multiple taxation, FIRS, internal revenue, tax burden |
| Business licensing and permits (j30c) | Russian Businesses In Rwanda Will Be Treated Same Way As Local Ones - Trade Minister - UrduPoint. | license, permit, business registration, bureaucracy, regulatory compliance, startup procedure, operating license, business permit, red tape, license approval, formalization, SME registration, licensing delays, regulatory authority, certificate of incorporation |
| Political instability (j30e) | Lesotho Holds Vote Six Months After Leader Fled Coup Attempt. | political unrest, instability, regime change, coup, protests, demonstration, civil unrest, election violence, political crisis, government collapse, military rule, transition, political risk, conflict, state failure, rebellion, power struggle |
| Corruption (j30f) | Corruption a major hindrance to Africa's industrialization: experts. | corruption, bribery, embezzlement, fraud, graft, kickbacks, anti-corruption, nepotism, money laundering, transparency, accountability, audit, public funds, illicit enrichment, ethics, scandal, whistleblower, procurement fraud |
| Labor regulations (l30a) | Fresh Hurdles Against Minimum Wage Takeoff. | labor law, minimum wage, employment regulation, collective bargaining, unions, labor dispute, hiring rules, work contract, termination, social protection, occupational safety, working hours, labor compliance, employment act, labor ministry |
| Inadequately educated workforce (l30b) | National Skills Competition launched in Accra. | skills gap, vocational training, education quality, technical education, TVET, unqualified workforce, human capital, school curriculum, workforce mismatch, employability, training center, youth unemployment, education reform, skill development, labor market needs |
| Electricity (c30a) | Zimbabwe secures more electricity imports to ease outages. | power supply, electricity, blackout, grid, power outage, load shedding, energy access, renewable energy, energy infrastructure, electricity distribution, energy sector, public utility, electrification, power generation, off-grid, rural electrification |
| Transport (d30a) | Railway Transport in South Africa 2017. LONDON, Oct. | infrastructure, road, transport, logistics, port, railway, airport, traffic congestion, public transport, transport costs, highway, freight, logistics corridor, bridge, transport network, inland transport, urban mobility |
| Customs and trade Regulations (d30b) | African Union Landmark Trade Pact Grows, as Benin, Nigeria Join. | customs, tariff, trade agreement, AfCFTA, import/export, trade facilitation, non-tariff barriers, cross-border trade, customs union, free trade zone, WTO, trade policy, clearance procedures, regional integration, value chains, customs duty, rules of origin |
| Access to finance (k30) | CIBN chief stresses importance of bank loans, economic growth. | credit, loan, interest rate, banking, financial inclusion, collateral, SME finance, microfinance, access to capital, commercial bank, credit constraint, debt, equity finance, lending, investment finance, financial services, creditworthiness |
| Access to land (g30a) | Historical land grievances at the Coast and elsewhere must be carefully settled. | land tenure, land rights, property registration, land dispute, title deed, land allocation, customary land, lease, expropriation, land reform, real estate, zoning, land market, land governance, land administration, land ownership |
| Crime, theft and disorder (i30) | Cattle Rustlers, Robbers & Kidnappers Paraded With Their Weapons. | crime, theft, robbery, kidnapping, armed violence, insecurity, gangs, public safety, law enforcement, criminal justice, militants, trafficking, bandits, street crime, police operation, vigilante, security challenge, cattle rustling, disorder |

Table A3 reports summary statistics for the variables employed in the research, the data source, and the measurement level (regional vs firm level).

**Table A3**
Summary Statistics Table

| | Obs | Mean | Std Dev | Source | Measurement Level |
|---|---|---|---|---|---|
| Foreign Ownership | 6232 | 0.104 | 0.266 | WBES | Firm |
| FDI Dummy | 6232 | 0.165 | 0.371 | WBES | Firm |
| Age (ln) | 6232 | 2.715 | 0.820 | WBES | Firm |
| Dummy Small | 6232 | 0.460 | 0.498 | WBES | Firm |
| Dummy Big | 6232 | 0.224 | 0.417 | WBES | Firm |
| Sales Growth | 6232 | 1.215 | 31.171 | WBES | Firm |
| Employment Growth | 6232 | 2.341 | 17.414 | WBES | Firm |
| Infrastructure | 6232 | 0.135 | 1.262 | WBES | Firm |
| Market Frictions | 6232 | 0.103 | 1.375 | WBES | Firm |
| Human Capital | 6232 | 0.008 | 1.206 | WBES | Firm |
| Governance | 6232 | 0.156 | 1.737 | WBES | Firm |
| Agglomeration | 6232 | 3.035 | 6.495 | WBES | Regional |
| Region Size | 6232 | 104.716 | 77.330 | WBES | Regional |

(*continued on next page*)

**Table A3** (*continued*)

| | Obs | Mean | Std Dev | Source | Measurement Level |
|---|---|---|---|---|---|
| SBS Courts (h30) | 6232 | 0.009 | 0.029 | ER | Regional |
| SBS Taxes (j30a and j30b) | 6232 | 0.005 | 0.021 | ER | Regional |
| SBS Business licensing and permits (j30c) | 6232 | 0.001 | 0.007 | ER | Regional |
| SBS Political instability (j30e) | 6232 | 0.060 | 0.154 | ER | Regional |
| SBS Corruption (j30f) | 6232 | 0.091 | 0.179 | ER | Regional |
| SBS Labor regulations (l30a) | 6232 | 0.001 | 0.005 | ER | Regional |
| SBS Inadequately educated workforce (l30b) | 6232 | 0.045 | 0.102 | ER | Regional |
| SBS Electricity (c30a) | 6232 | 0.045 | 0.117 | ER | Regional |
| SBS Transport (d30a) | 6232 | 0.040 | 0.101 | ER | Regional |
| SBS Customs and trade Regulations (d30b) | 6232 | 0.019 | 0.082 | ER | Regional |
| SBS Access to finance (k30) | 6232 | 0.075 | 0.154 | ER | Regional |
| SBS Access to land (g30a) | 6232 | 0.021 | 0.070 | ER | Regional |
| SBS Crime, theft and disorder (i30) | 6232 | 0.002 | 0.006 | ER | Regional |
| Sentiment | 6232 | 0.002 | 0.038 | ER | Regional |
| N Docs | 6232 | 18.160 | 45.004 | ER | Regional |
| Nightlight Activity | 6232 | 1.835 | 3.964 | VNL | Regional |
| Country Coverage | 6232 | 0.037 | 0.054 | ER | Regional |

Table A4 reports the list of countries analyzed, the number of firms, and each country's foreign ownership (average value).

**Table A4**
Total number of firms and average foreign ownership by country

| Country | Total no. of firms | Foreign ownership |
|---|---|---|
| Benin | 67 | 0.147 |
| Burundi | 48 | 0.205 |
| Cameroon | 99 | 0.078 |
| Chad | 69 | 0.106 |
| Cote d'Ivoire | 104 | 0.230 |
| Egypt | 1163 | 0.051 |
| Eswatini | 69 | 0.098 |
| Ethiopia | 379 | 0.126 |
| Gambia | 72 | 0.110 |
| Guinea | 21 | 0.048 |
| Kenya | 451 | 0.123 |
| Lesotho | 74 | 0.166 |
| Liberia | 67 | 0.242 |
| Malawi | 181 | 0.214 |
| Mali | 93 | 0.198 |
| Mauritania | 48 | 0.140 |
| Morocco | 461 | 0.089 |
| Mozambique | 150 | 0.194 |
| Namibia | 122 | 0.056 |
| Niger | 34 | 0.092 |
| Nigeria | 1374 | 0.046 |
| Rwanda | 116 | 0.130 |
| Senegal | 245 | 0.080 |
| Sierra Leone | 76 | 0.068 |
| South Sudan | 75 | 0.479 |
| Sudan | 77 | 0.019 |
| Togo | 34 | 0.606 |
| Zambia | 176 | 0.273 |
| Zimbabwe | 287 | 0.100 |
| Total | 6232 | 0.104 |

Table A5 reports the list of regions analyzed and the number of firms. Each region is defined by the World Bank Enterprise Surveys. We have a total of 107 regions in our dataset.

**Table A5**
List of Regions

| Region | N Firms | Region | N Firms | Region | N Firms |
|---|---|---|---|---|---|
| Abia | 61 | Kebbi | 71 | Nimba | 8 |
| Abidjan | 80 | Kenema | 14 | Northern Upper Egypt | 146 |

(*continued on next page*)

**Table A5** (*continued*)

| Region | N Firms | Region | N Firms | Region | N Firms |
|---|---|---|---|---|---|
| Abuja | 78 | Khartoum | 32 | Nouadhibou | 23 |
| Addisababa | 181 | Kiambu | 71 | Nouakchott | 25 |
| Amhara | 21 | Kigali | 41 | Ogun | 77 |
| Anambra | 62 | Kilifi | 7 | Oriental | 40 |
| Atlantique, Borgou, Mono, Ouémé | 15 | Kirinyaga | 40 | Oromia | 66 |
| Bahri | 15 | Kisumu | 17 | Oshakati/ Ondangwa/ Ongwediva | 24 |
| Bamako | 60 | Kitwe | 32 | Oyo | 44 |
| Blantyre | 99 | Kwara | 76 | Rabat-Salé-Kénitra | 69 |
| Bo | 13 | Lagos | 152 | Rest of the Country | 24 |
| Bombali | 10 | Lesotho | 74 | Saint-Louis | 18 |
| Bujumbura | 48 | Lilongwe | 63 | SNNPR | 28 |
| Bulawayo | 92 | Littoral | 91 | Sofala | 52 |
| Béni Mellal-Khénifra and Drâa-Tafilalet | 8 | Livingstone | 17 | Sokoto | 70 |
| Cabo Delgado | 10 | Lomé | 34 | Souss-Massa | 18 |
| Casablanca-Settat | 81 | Lusaka | 97 | Southern Province | 41 |
| Center | 40 | Machakos | 24 | Southern Upper Egypt | 91 |
| Conakry | 21 | Manica | 16 | Suez Region | 81 |
| Cross River | 53 | Manicaland | 36 | Tanger-Tétouan-Al Hoceima | 97 |
| Dakar | 157 | Margibi | 15 | Tete | 14 |
| Dredawa | 19 | Marrakech-Safi | 62 | Thiès | 37 |
| Enugu | 64 | Middle and East Delta | 268 | Tigray | 64 |
| Eswatini | 69 | Midlands | 32 | Trans Nzoia | 10 |
| Frontier | 52 | Mombasa | 44 | Uasin Gishu | 26 |
| Fès-Meknès | 86 | Montserrado | 44 | Unknown | 30 |
| Gambia | 72 | Mopti, Ségou, Sikasso | 33 | West | 20 |
| Gombe | 62 | Mzimba | 9 | West Delta | 208 |
| Greater Cairo | 317 | N'Djamena | 69 | Western Province | 34 |
| Harare | 127 | Nairobi | 174 | Western Urban | 39 |
| Jigawa | 65 | Nakuru | 38 | Windhoek/ Okahandja | 98 |
| Juba | 70 | Nampula | 47 | Yei | 5 |
| Kaduna | 60 | Nasarawa | 81 | Zambézia | 11 |
| Kano | 87 | Ndola | 30 | Zamfara | 63 |
| Kaolack | 33 | Niamey | 34 | Zomba | 10 |
| Katsina | 69 | Niger | 79 | | |

Table A6 Panel A reports the full specification of the 2SLS regressions used to estimate regression coefficients in Fig. 3. Panel B decomposes the PCA coefficients into their single components.

**Table A6**
Panel A shows the interaction results used to obtain the regression coefficients reported in Fig. 3. Panel B reports the results using the variables used to build the PCA factors for Infrastructure and Human Capital, respectively

| | Panel (A) | |
|---|---|---|
| | Dependent Variable: % of Foreign Ownership | |
| Infrastructure # SBS Justice access | -0.898*** | |
| | (0.163) | |
| Infrastructure | 0.031 | 0.022** |
| | (0.028) | (0.010) |
| SBS Justice access | 0.961*** | 0.757*** |
| | (0.160) | (0.179) |
| Human Capital # SBS Justice access | | -0.845*** |
| | | (0.220) |
| Human Capital | -0.003 | 0.017 |
| | (0.010) | (0.027) |
| Sentiment | 0.303* | 0.329* |
| | (0.177) | (0.167) |
| Age (Ln) | -0.010 | -0.006 |
| | (0.010) | (0.011) |
| Dummy Small Size | 0.089*** | 0.091*** |
| | (0.022) | (0.022) |
| Dummy Big Size | 0.217*** | 0.213*** |
| | (0.041) | (0.043) |
| Agglomeration | 0.000*** | 0.000*** |
| | (0.000) | (0.000) |
| Region Size | -0.000** | -0.000* |

(continued on next page)

**Table A6** (*continued*)

| | Panel (A) | |
|---|---|---|
| | Dependent Variable: % of Foreign Ownership | |
| | (0.000) | (0.000) |
| Sales Growth | -0.000 | -0.000 |
| | (0.000) | (0.000) |
| Employment Growth | -0.000 | -0.000 |
| | (0.001) | (0.001) |
| N Docs | 0.000*** | 0.001*** |
| | (0.000) | (0.000) |
| Nightlight Activity | 0.003** | 0.003** |
| | (0.001) | (0.001) |
| Country Coverage | 0.199 | 0.260 |
| | (0.296) | (0.316) |
| Market Frictions | -0.015 | -0.015 |
| | (0.010) | (0.012) |
| Governance | -0.000 | -0.004 |
| | (0.009) | (0.009) |
| Observations | 5902 | 5902 |
| Country-industry f.e. | Yes | Yes |
| R Squared | 0.065 | 0.052 |
| N Countries | 29 | 29 |
| N Regions | 107 | 107 |

| | Panel (B) | | | | |
|---|---|---|---|---|---|
| | Infrastructure | | | Human Capital. | |
| | (1) | (2) | (3) | (4) | (5) |
| c30a # SBS Justice access | -1.045*** | | | | |
| | (0.297) | | | | |
| c30a | 0.048 | | | | |
| | (0.032) | | | | |
| SBS Justice access | 3.317*** | 2.199*** | 1.039*** | 1.545*** | 1.110*** |
| | (0.790) | (0.455) | (0.354) | (0.402) | (0.343) |
| d30a # SBS Justice access | | -0.995*** | | | |
| | | (0.212) | | | |
| d30a | | 0.030** | | | |
| | | (0.015) | | | |
| d30b # SBS Justice access | | | -0.637*** | | |
| | | | (0.236) | | |
| d30b | | | -0.012 | | |
| | | | (0.030) | | |
| l30a # SBS Justice access | | | | -0.810*** | |
| | | | | (0.232) | |
| l30a | | | | 0.018 | |
| | | | | (0.040) | |
| l30b # SBS Justice access | | | | | -0.502** |
| | | | | | (0.194) |
| l30b | | | | | 0.010 |
| | | | | | (0.025) |
| Sentiment | 0.226 | 0.309* | 0.379** | 0.319* | 0.329** |
| | (0.206) | (0.178) | (0.174) | (0.166) | (0.162) |
| Age (Ln) | -0.008 | -0.011 | -0.009 | -0.007 | -0.007 |
| | (0.010) | (0.010) | (0.011) | (0.010) | (0.010) |
| Dummy Small Size | 0.086*** | 0.091*** | 0.101*** | 0.091*** | 0.090*** |
| | (0.021) | (0.021) | (0.024) | (0.022) | (0.022) |
| Dummy Big Size | 0.217*** | 0.220*** | 0.230*** | 0.213*** | 0.215*** |
| | (0.040) | (0.043) | (0.044) | (0.043) | (0.042) |
| Agglomeration | 0.000*** | 0.000*** | 0.000*** | 0.000*** | 0.000*** |
| | (0.000) | (0.000) | (0.000) | (0.000) | (0.000) |
| Region Size | -0.000* | -0.000** | -0.000** | -0.000 | -0.000** |
| | (0.000) | (0.000) | (0.000) | (0.000) | (0.000) |
| Sales Growth | 0.000 | -0.000 | 0.000 | -0.000 | 0.000 |
| | (0.000) | (0.000) | (0.000) | (0.000) | (0.000) |
| Employment Growth | -0.000 | -0.000 | -0.000 | -0.000 | -0.000 |
| | (0.001) | (0.001) | (0.001) | (0.001) | (0.001) |
| N Docs | 0.000*** | 0.001*** | 0.001*** | 0.001*** | 0.001*** |
| | (0.000) | (0.000) | (0.000) | (0.000) | (0.000) |
| Nightlight Activity | 0.003** | 0.004** | 0.004** | 0.003** | 0.003** |
| | (0.001) | (0.002) | (0.002) | (0.001) | (0.001) |
| Country Coverage | 0.193 | 0.183 | 0.223 | 0.286 | 0.188 |
| | (0.325) | (0.314) | (0.305) | (0.345) | (0.291) |
| Human Capital | -0.002 | -0.002 | -0.000 | | |

(continued on next page)

**Table A6** (*continued*)

| | | | | | |
|---|---|---|---|---|---|
| | (0.009) | (0.010) | (0.009) | | |
| Market Frictions | -0.018* | -0.014 | -0.010 | -0.017* | -0.017 |
| | (0.009) | (0.010) | (0.009) | (0.010) | (0.011) |
| Governance | 0.001 | 0.000 | 0.008 | -0.004 | -0.003 |
| | (0.007) | (0.007) | (0.009) | (0.010) | (0.007) |
| Infrastructure | | | | 0.022** | 0.021* |
| | | | | (0.010) | (0.011) |
| Observations | 5902 | 5902 | 5902 | 5902 | 5902 |
| Country-industry f.e. | Yes | Yes | Yes | Yes | Yes |
| R Squared | 0.052 | 0.049 | 0.047 | 0.055 | 0.067 |
| N Countries | 29 | 29 | 29 | 29 | 29 |
| N Regions | 107 | 107 | 107 | 107 | 107 |

Table A7 presents the results using a foreign dummy variable as the dependent variable in columns 5 through 8. In columns 1 through 4, however, the percentage of foreign ownership is used as the dependent variable, limiting the sample to companies with a positive value for that variable.

**Table A7**
Intensive vs Extensive margin

| | **Intensive Margin** | | | | **Extensive Margin** | | | |
|---|---|---|---|---|---|---|---|---|
| | (1) | (2) | (3) | (4) | (5) | (6) | (7) | (8) |
| SBS Access to land (g30a) | 0.457* | | | 0.519* | 0.193 | | | 0.059 |
| | (0.235) | | | (0.269) | (0.131) | | | (0.095) |
| SBS Crime, theft and disorder (i30) | | -3.795 | | -4.535 | | -4.193*** | | -3.158*** |
| | | (2.502) | | (3.159) | | (1.202) | | (1.132) |
| SBS Inadequately educated workforce (l30b) | | | -0.110 | -0.160 | | | 0.343*** | 0.292*** |
| | | | (0.249) | (0.235) | | | (0.085) | (0.092) |
| Controls | YES | YES | YES | YES | YES | YES | YES | YES |
| Country x Industry FE | YES | YES | YES | YES | YES | YES | YES | YES |
| Observations | 1122 | 1122 | 1122 | 1122 | 6232 | 6232 | 6232 | 6232 |
| R Squared | 0.571 | 0.570 | 0.570 | 0.572 | 0.372 | 0.373 | 0.373 | 0.374 |

How a location is portrayed in the news media landscape can also be operationalized using alternative approaches to the SBS. In Table A8 we compare the explanatory power of SBS with that of article volume, keyword frequency, and sentiment. We measure article volume as the number of articles related to a specific region-theme that have been published. Keyword frequency measures the frequency of keywords related to those themes within the articles. Each panel in Table A8 focuses on a different theme (access to land in Panel A, crime, theft and disorder in Panel B, and inadequately educated workforce in Panel C).

**Table A8**
Comparison between SBS and other textual metrics

| Panel A. SBS Access to land | | | | | | |
|---|---|---|---|---|---|---|
| | (1) Baseline | (2) SBS | (3) Sentiment | (4) Article count | (5) Keyword Frequency | (6) All measures |
| SBS | | 0.190** | | | | 0.296* |
| | | (0.073) | | | | (0.154) |
| Sentiment | | | 0.274* | | | 0.149 |
| | | | (0.158) | | | (0.168) |
| Article count | | | | -0.010* | | -0.008 |
| | | | | (0.006) | | (0.007) |
| Keyword Frequency | | | | | -0.000 | -0.001 |
| | | | | | (0.001) | (0.001) |
| Observations | 6232 | 6232 | 6232 | 6232 | 6232 | 6232 |
| R2 / Delta vs. baseline (p.p.) | 0.3818 | +0.03 | +0.02 | +0.03 | +0.00 | +0.09 |
| Controls | Yes | Yes | Yes | Yes | Yes | Yes |
| Country x industry FE | Yes | Yes | Yes | Yes | Yes | Yes |
| Panel B. SBS Crime, theft and disorder | | | | | | |
| | (1) Baseline | (2) SBS | (3) Sentiment | (4) Article count | (5) Keyword Frequency | (6) All measures |
| SBS | | -4.500*** | | | | -2.425** |
| | | (1.070) | | | | (1.099) |
| Sentiment | | | 0.274* | | | 0.147 |

(continued on next page)

**Table A8** (*continued*)

| | | | | | | |
|---|---|---|---|---|---|---|
| | | | (0.158) | | | (0.155) |
| Article count | | | | -0.086*** | | -0.055* |
| | | | | (0.025) | | (0.030) |
| Keyword Frequency | | | | | -0.018*** | -0.003 |
| | | | | | (0.005) | (0.005) |
| Observations | 6232 | 6232 | 6232 | 6232 | 6232 | 6232 |
| R2 / Delta vs. baseline (p.p.) | 0.3818 | +0.12 | +0.02 | +0.17 | +0.12 | +0.20 |
| Controls | Yes | Yes | Yes | Yes | Yes | Yes |
| Country x industry FE | Yes | Yes | Yes | Yes | Yes | Yes |
| Panel C. SBS Inadequately educated workforce | | | | | | |
| | (1) | (2) | (3) | (4) | (5) | (6) |
| | Baseline | SBS | Sentiment | Article count | Keyword Frequency | All measures |
| SBS | | 0.169*** | | | | 0.208** |
| | | (0.054) | | | | (0.091) |
| Sentiment | | | 0.274* | | | 0.357** |
| | | | (0.158) | | | (0.151) |
| Article count | | | | 0.003 | | -0.001 |
| | | | | (0.009) | | (0.010) |
| Keyword Frequency | | | | | 0.001 | -0.000 |
| | | | | | (0.000) | (0.001) |
| Observations | 6232 | 6232 | 6232 | 6232 | 6232 | 6232 |
| R2 / Delta vs. baseline (p.p.) | 0.3818 | +0.05 | +0.02 | +0.00 | +0.01 | +0.08 |
| Controls | Yes | Yes | Yes | Yes | Yes | Yes |
| Country x industry FE | Yes | Yes | Yes | Yes | Yes | Yes |

While the increase in model fit produced by each media measure is modest, SBS generates the largest increase in $R^2$ for inadequately educated workforce and an increase comparable to article count for access to land. Article count yields a larger increase for crime, theft and disorder. When all media measures capturing the importance of a theme (article volume, keyword frequency, and SBS) are included jointly, SBS remains statistically significant across all themes analyzed.

In addition to comparing the explanatory power of SBS with that of alternative measures, in Table A9 we break down SBS into its three components to better understand which of the three drives our results. Table A9 shows the regression coefficients of the individual SBS components for each theme. We use prevalence in Column 1, diversity in Column 2, and connectivity in the last column. In each specification, the dependent variable remains the percentage of foreign ownership, as in the main analyses.

**Table A9**
Single SBS components

| | Dependent Variable: % of Foreign Ownership | | |
|---|---|---|---|
| | (1) **Prevalence** | (2) **Diversity** | (3) **Connectivity** |
| SBS Access to land (g30a) | 0.113** | 0.141*** | 0.097** |
| | (0.055) | (0.053) | (0.040) |
| SBS Crime, theft and disorder (i30) | -5.442*** | -3.159*** | -3.533*** |
| | (1.365) | (0.747) | (0.913) |
| SBS Inadequately educated workforce (l30b) | 0.136** | 0.109 | 0.110* |
| | (0.061) | (0.067) | (0.059) |
| Observations | 6232 | 6232 | 6232 |
| R-squared | 0.3836 | 0.3835 | 0.3832 |
| Controls | Yes | Yes | Yes |
| Country x industry FE | Yes | Yes | Yes |

The coefficients of the individual components are consistent with the SBS coefficients in Table 4. When the three SBS components are used separately as explanatory variables, their coefficients generally have the same sign as the corresponding SBS coefficient, indicating that prevalence, diversity, and connectivity point in the same direction in their relationship with foreign ownership. This pattern is consistent with the theoretical foundation of the SBS, according to which the three components represent complementary dimensions of the same underlying construct: the semantic importance of a theme within a text (Fronzetti Colladon, 2018). However, the statistical significance of the individual components varies across media frames. For access to land, diversity shows the strongest and most statistically significant relationship. For crime, theft and disorder, no substantial differences emerge across the three components, as all three are statistically significant at the 1% level. For inadequately educated workforce, prevalence shows the strongest and most statistically significant relationship among the three components. This decomposition is intended as a purely illustrative exercise, as our choice to use the SBS is primarily grounded in the theoretical arguments discussed in Section 2.1 rather than in the empirical performance of its individual components.

Figure A1 plots the marginal effects of infrastructure and human capital on foreign ownership across values of SBS justice access. In both panels, the dashed vertical line indicates the level of SBS beyond which the PCA components become significant.

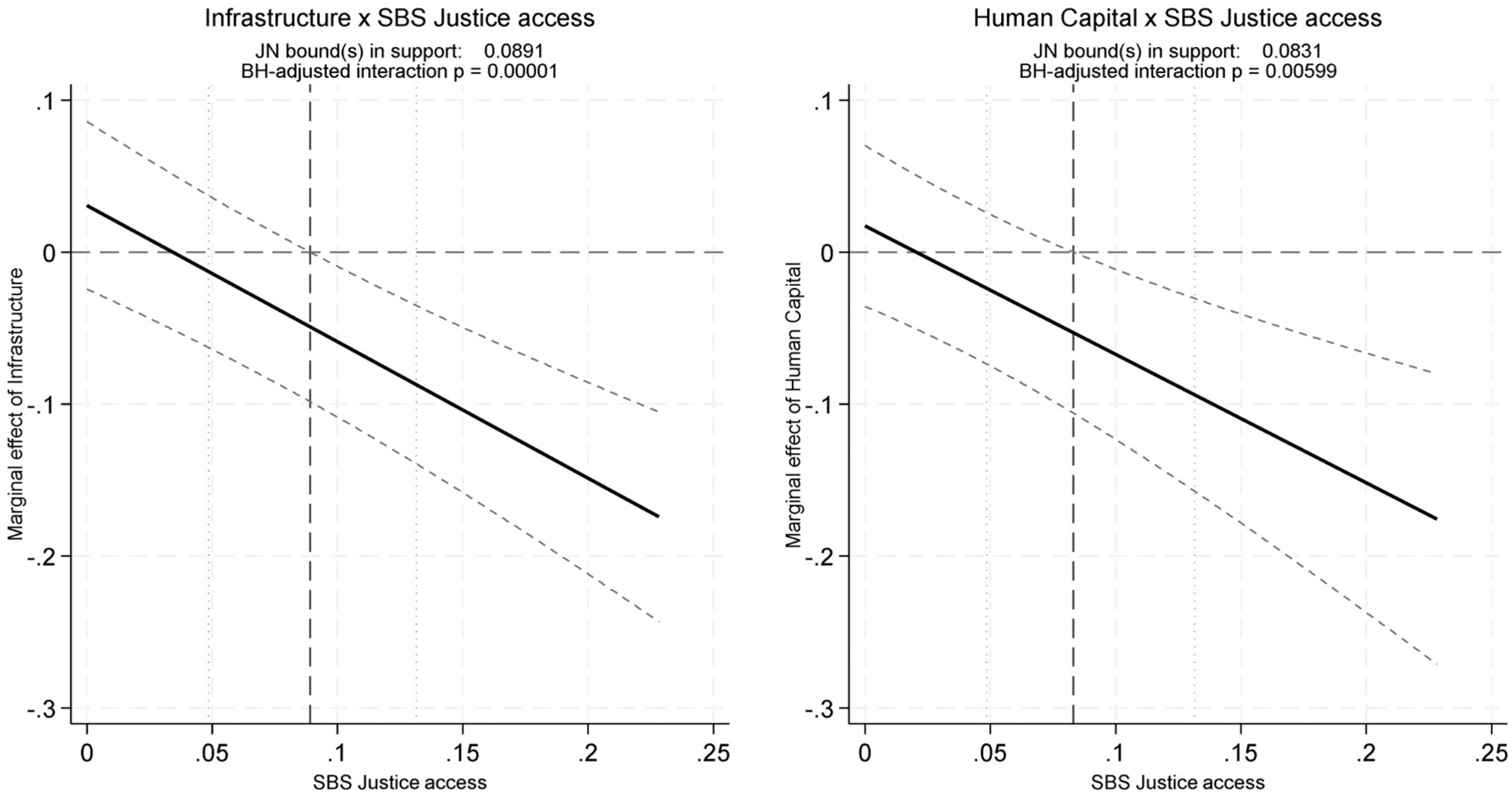

**Fig. A1.** Johnson-Neyman Analysis

Table A10 presents the same model specification as in Table 2, but using lagged media variables. The results are consistent with Table 3; the direction of the coefficients for the SBS variables remains positive and statistically significant. Compared to the IV specification, the coefficient for inadequately educated workforce also becomes statistically significant.

**Table A10**
Regression results using lagged media variable

| | Dependent Variable: % of Foreign Ownership | | | |
|---|---|---|---|---|
| | (1) | (2) | (3) | (4) |
| SBS Access to land (g30a) | 0.217** | | | 0.146* |
| | (0.099) | | | (0.076) |
| SBS Crime, theft and disorder (i30) | | 2.056*** | | 1.850*** |
| | | (0.505) | | (0.473) |
| SBS Inadequately educated workforce (l30b) | | | 0.215*** | 0.228*** |
| | | | (0.080) | (0.080) |
| Observations | 4062 | 4062 | 4062 | 4062 |
| R Squared | 0.40 | 0.40 | 0.40 | 0.40 |
| Controls | Yes | Yes | Yes | Yes |
| N Countries | 21 | 21 | 21 | 21 |
| N Regions | 70 | 70 | 70 | 70 |

Table A11 presents the results of the Principal Component Analysis. The first column lists the name of the component used in the paper, and the second column lists the business obstacles that comprise it. The last three columns list, respectively, the loading of each variable, the eigenvalue of the factor, and the explained variance.

**Table A11**
Results of the PCA component

| Investment-climate index | Business obstacle | PC1 loading | Eigenvalue | Variance explained |
|---|---|---|---|---|
| Governance | Tax rates | 0.757 | 3.077 | 51.29% |
| | Tax administration | 0.755 | | |
| | Business licensing and permits | 0.712 | | |
| | Political instability | 0.706 | | |
| | Corruption | 0.722 | | |
| | Courts | 0.638 | | |

(*continued on next page*)

**Table A11** (*continued*)

| Investment-climate index | Business obstacle | PC1 loading | Eigenvalue | Variance explained |
|---|---|---|---|---|
| Infrastructure | Electricity | 0.646 | 1.606 | 53.54% |
| | Transport | 0.822 | | |
| | Customs and trade regulations | 0.717 | | |
| Human Capital | Labor regulations | 0.865 | 1.497 | 74.83% |
| | Inadequately educated workforce | 0.865 | | |
| Market Frictions | Practices of competitors in the informal sector | 0.604 | 1.910 | 47.75% |
| | Access to finance | 0.714 | | |
| | Access to land | 0.730 | | |
| | Crime, theft, and disorder | 0.708 | | |

*Themes' Content Summary*

To address concerns about the potential abstract nature of media-derived variables (SBS media importance scores of each theme), we generated concise, human-interpretable summaries of each theme using a large language model (ChatGPT version 4.0, accessed on October 2nd 2025, using temperature equal to 0.7) applied to a sample of news excerpts. For each theme reported in Table A2, we selected ten articles from the overall corpus in which the theme was most semantically relevant, as measured by the SBS. The choice of ten articles was guided by the input constraints of the language model; using slightly more or fewer articles produced very similar outcomes. Since this procedure served only as an auxiliary step to improve the interpretability of the media-derived variables, rather than their computation, the precise number of articles summarized is not critical to our analysis. We then submitted the following standardized prompt to ChatGPT (we define this prompt as "Summary Prompt" in the rest of the manuscript):

**Prompt Example:**

*I am analyzing a corpus of news to measure the semantic importance of the theme* [Theme of interest]. *Below are 10 short excerpts from news articles where this topic is prominently discussed. Please write a concise summary explaining how this theme is represented in media discourse. The goal is to provide a clear, grounded interpretation of what* [Theme of interest] *refers to in these articles*

*News Excerpts:*
[10 text snippets here]

**Fig. A2.** Input prompt to ChatGPT used to summarize the content of each theme

Because ChatGPT can occasionally produce inaccurate interpretations of textual input, we implemented an additional validation step to ensure the reliability of our media-derived variables. Specifically, we sought to confirm that texts receiving high SBS scores for a given variable truly centered on the intended theme. To accomplish this, we used the following prompt (we define this prompt as "Identification Prompt" in the rest of the manuscript):

**Prompt Example:**

*Read all of the following news excerpts. Identify the single overarching theme that best captures them all. Respond in no more than three words, without any explanations.*

*News Excerpts:*

[10 text snippets here]

**Fig. A3.** Input prompt to ChatGPT used to identify the main theme of each group of news excerpts

In this validation task, rather than constraining the model to predefined themes, we allowed it to independently determine the most salient topic and express it concisely. If the model produced keywords that diverged from the conceptual construct we intended to measure, this would indicate that our scoring method was capturing irrelevant content. However, as shown in our results, the model's responses consistently aligned with the theoretical definitions of our variables, thereby supporting the validity of our media-derived measures.

The LLM-generated answers for each prompt are as follows:

Courts(h30)

*Summary Prompt.* This theme reflects how the judicial system is portrayed in news reporting, with a focus on legal infrastructure, judicial reforms, and access to justice. Coverage typically addresses:

- Judicial modernization, such as laws targeting faster criminal proceedings and curbing procedural delays (e.g., the Administration of Criminal Justice Act abolishing stay of proceedings).
- Judicial independence, particularly in the context of state-level financial autonomy, where commentators stress reduced executive influence.
- Systemic inefficiencies, notably the slow pace of high-profile corruption trials and the structural backlog of unresolved cases.
- Barriers to access, especially for the poor or vulnerable, as noted in African regional discussions on legal aid and community justice institutions.
- Sector-specific dispute resolution, where regulatory bodies (e.g., NCC, NITDA) emphasize legal mechanisms to address governance and compliance issues in technology and telecom sectors.
- Perceptions of corruption, with some sources pointing to mistrust in the judiciary's neutrality, especially in arbitration or politically exposed cases.

Example Phrasing from Articles:

- "Unending trial of politically exposed persons in corruption cases"
- "Access to justice remains a work in progress in Africa"
- "Judiciary's financial autonomy a long-awaited development"

*Identification Prompt.* Legal and Regulatory Reform, Governance and Regulation, Regulation and Justice.

Taxes (j30a and j30b)

*Summary Prompt.* This theme captures media discussion around the structure, enforcement, and implications of taxation policy. It reflects both national and regional developments, and typically involves the following dimensions:

- Tax burden and competitiveness: Reports frequently highlight comparative tax rates (e.g., Kenya's gambling tax being the highest in Africa) and their implications for investment, particularly in sensitive or high-growth sectors like gaming.
- Tax system inefficiencies: Articles emphasize systemic problems such as fragmented databases, inconsistent application across tiers of government, and poor compliance enforcement, particularly in Nigeria.
- Multiple and informal taxes: Business leaders and analysts express concern over overlapping levies, unpredictable charges, and harassment by tax authorities, which are seen as disincentives to local and foreign investment.
- Tax reform and policy development: Discussions include ongoing tax reform efforts (e.g., the National Tax Policy, Voluntary Asset Declaration Scheme, and security-related levies) aimed at expanding the tax base and reducing dependence on oil revenues.
- Enforcement and compliance: Commentary also touches on the aggressive tone of enforcement measures — including threats of prosecution for tax evasion — and the political sensitivity surrounding high-profile defaulters.
- Civic responsibility and leadership: Some articles call attention to ethical leadership, where nominees for public office face scrutiny over their personal tax compliance.

Example Phrasing from Articles:

- "Multiple taxations…are discouraging investment in the country."
- "Nigeria has not derived the maximum benefits of its tax system in terms of revenue generation."
- "Kenya's betting tax is the highest in the region… ahead of Las Vegas."
- "FIRS has bared its fangs against tax evaders."

*Identification Prompt.* Taxation Challenges, Tax system challenges, Taxation Challenges.

Business Licensing and Permits (j30c)

*Summary Prompt.* This theme captures how media reports on the regulatory environment for establishing and operating a business. It includes licensing requirements, permitting procedures, administrative efficiency, and institutional reforms. Key elements from the coverage include:

- Ease of doing business: Rwanda is held up as a regional benchmark for streamlined licensing, where businesses can be registered within hours via one-stop centers, reducing bureaucratic friction.
- Cost of entry: Articles discuss specific fee structures for different business sizes and sectors (e.g., education, manufacturing, hospitality) in counties like Nakuru and cities like Nairobi. These licensing fees are often scaled to business size or employee count.

- Policy inconsistency and regulatory overreach: In Nigeria, concerns arise over proposed monopolistic regulations, such as the NIHOTOURS bill, which would centralize control of the tourism licensing regime — a move criticized as harmful to industry development.
- Local content requirements: In Ghana and Nigeria, policies linking business licensing to local participation, especially in oil, gas, and mining sectors, are highlighted. These policies incentivize partnerships with domestic firms and promote national workforce development.
- Incentive regimes: The Nigerian government, for instance, offers tax holidays and permit-related waivers to attract investors in targeted sectors like mining, showcasing how permitting frameworks are used to shape investment flows.
- Administrative bottlenecks: Commentary warns that poor licensing practices and excessive red tape (especially at the sub-national level) can deter investment, despite broader economic promotion efforts.

Example Phrasing from Articles:

- "It is very easy to start a company. It takes roughly six hours."
- "Bureaucratic system that makes it easier for a camel to pass through the eye of a needle than for an investor…"
- "Proposed bill… described as the death knell of hospitality and tourism."
- "Tax holidays and duty waivers on imported mining equipment."

*Identification Prompt.* Investment and Development, Business investment climate, Business climate investments.

Political Instability (j30e)

*Summary Prompt.* This theme encompasses news coverage of political uncertainty, electoral violence, weak governance, and the erosion of state authority — especially as these affect economic performance and public trust. Media references to political instability typically include:

- Electoral crises and contested transitions: Articles describe allegations of vote rigging, military interference in democratic processes, and election-related violence (e.g., Lesotho coup attempt, Tanzania vote controversy, Zambia's violent campaign, Nigeria's election glitches and security fears).
- Armed conflict and insurgency: Stories from South Sudan and the Niger Delta highlight how rebellion, civil war, and militia activities threaten not only human security but also strategic economic sectors (e.g., oil production).
- Governance failure and institutional paralysis: Reports on Libya and Nigeria underscore how prolonged political dysfunction (e.g., policy reversals, lack of continuity, conflict between executive and legislative branches) hampers economic management and service delivery.
- Public distrust and social unrest: Several pieces cite declining public confidence in leadership, citing hunger, inflation, and unmet development expectations as root causes of popular discontent, particularly in Nigeria and the Niger Delta.
- Security policy and national cohesion: Commentary suggests that lack of comprehensive security planning and overreliance on military solutions (e.g., South Sudan, Nigeria) further exacerbates fragility.

Example Phrasing from Articles:

- "Political turmoil has hobbled the government's ability to govern."
- "Zambians go to the polls… amid a backdrop of a tanking economy."
- "We are in deep trouble… hunger everywhere… conflict between politics and governance."
- "South Sudan's rebels say they seize oil hub Bentiu."
- "Lesotho holds vote six months after leader fled coup attempt."

*Identification Prompt.* Political instability, Political instability, Political instability.

Corruption (j30f)

*Summary Prompt.* This theme encompasses public sector and political corruption, including misuse of public office, bribery, embezzlement, patronage, and opaque governance practices. Media references to corruption frequently highlight its systemic nature, link it to economic underdevelopment, and reflect both allegations and institutional responses.

Contextual Examples from News Coverage:

- Allegations Involving Political Leaders: Reports detail high-level accusations, such as the U.S. claim that Chad's president accepted a $2 million bribe for oil rights. These stories highlight the cross-border dimensions of corruption involving international firms and national elites.
- Institutional Corruption and Developmental Impact: Articles underscore how corruption impedes industrialization and economic governance in Africa. Experts argue that entrenched graft undermines state capacity and distorts policy implementation in key sectors like infrastructure and energy.
- Public Protests and Political Response: Media frequently covers anti-corruption protests by labor unions and civil society, with calls for stricter penalties (e.g., capital punishment in Nigeria) and judicial reforms to prevent obstruction of prosecutions.

- Governance within Political Parties: The internal handling of corruption within ruling parties is a recurring topic (e.g., calls within South Africa's ANC to expel corrupt members), pointing to concerns about institutional accountability and reputational damage.
- Corruption in Revenue and Local Administration: Coverage of regional governance includes examples of financial mismanagement (e.g., revenue leakages in Trans Nzoia County), with proposed reforms such as privatized collection systems sparking debate over transparency and trust.
- Historical and Cultural Narratives: Some media stories place corruption within broader political histories (e.g., Chicago's 19th-century underworld or critiques of elite-driven institutions), framing corruption as both a structural and moral issue affecting governance worldwide.

*Identification Prompt.* Corruption crisis, Widespread government corruption, Corruption Crisis.

Labor Regulations (l30a)

*Summary Prompt.* This theme captures discussions around the governance, enforcement, and contestation of rules and policies regulating employment relations, workplace standards, and union activities. In media discourse, this includes public debates over wage laws, labor strikes, employment conditions, ministerial actions, enforcement mechanisms, union demands, and tensions between workers and state authorities.

Contextual Examples from News Coverage:

- Ministerial Leadership: Reports of political appointments (e.g., Chris Ngige's reappointment) often reflect how ministerial leadership is perceived in shaping labor policy and regulatory enforcement.
- Wage Regulations: Coverage includes challenges in implementing new national minimum wage laws, particularly tensions between federal authorities and labor unions regarding fair salary structures.
- Workplace Conditions and Workers' Rights: Articles highlight grievances such as unpaid salaries, job insecurity, casualization, and threats of punitive actions against striking workers, indicating friction between regulatory protections and actual labor practices.
- Union Activities and Training: Media also tracks union capacity-building efforts and strategic responses to economic downturns, emphasizing the evolving regulatory challenges unions face.
- Sector-Specific Regulations: Industry-specific labor rules (e.g., in dairy or steel sectors) are discussed in terms of how they affect small producers and compliance burdens, often framed as part of broader regulatory or economic reform debates.
- Legal Disputes Over Strikes: Reports reflect on the legality of industrial actions and the use of courts to resolve or suppress labor disputes, illustrating how labor regulation is contested in practice.

*Identification Prompt.* Labour relations struggles, Labour and Employment, Labour and Regulation.

Inadequately Educated Workforce (l30b)

*Summary Prompt.* This theme refers to the mismatch between the skills, qualifications, and competencies of a country's labor force and the demands of its economy. It includes systemic weaknesses in education and training systems, poor teacher quality, outdated curricula, lack of technical and vocational training, and employer dissatisfaction with graduate capabilities.

Contextual Examples from News Coverage:

- Skills Gaps and Unemployment: Media articles frequently link youth unemployment to a shortage of job-relevant skills. Stakeholders point to industries that cannot find adequately trained personnel even as job seekers remain unemployed, reflecting a major disconnect between education systems and labor market needs.
- Emphasis on Technical and Vocational Training (TVET): National efforts to revamp vocational education (e.g., Ghana's National Skills Competition or China-Kenya vocational partnerships) are presented as targeted responses to the skills gap. These initiatives aim to foster employability, especially in manufacturing, agriculture, and green energy sectors.
- Teacher Quality Deficiencies: Several reports identify unqualified teachers as a central cause of poor learning outcomes. For example, Niger State's admission that 60% of its public school teachers are unqualified underscores institutional failures in ensuring a competent teaching workforce.
- Institutional and Infrastructure Challenges: Articles describe the physical and academic deterioration of technical institutions, including outdated programs, overstretched facilities, and misaligned course offerings. These factors perpetuate a cycle of underprepared graduates entering the workforce.
- Calls for Curriculum Reform: Education experts argue that obsolete curricula have left graduates ill-equipped for contemporary job markets, calling for urgent updates to reflect global standards, research, and evolving economic priorities.
- Social Attitudes Toward Educators: Critical commentary, such as warnings against treating teachers "like Libyan slaves," highlight how the status and welfare of educators affect the broader health of the workforce pipeline.

*Identification Prompt.* Education and Skills, Skills-based education reform, Skills-based education.

Electricity (c30a)

*Summary Prompt.* This theme encompasses the availability, reliability, and governance of electricity supply, including issues such as generation shortfalls, infrastructure failures, policy responses, tariff regulation, off-grid initiatives, and foreign investment in power systems. Media discussion reflects electricity as both a constraint on and an enabler of economic development and social well-being.

Contextual Examples from News Coverage:

- Power Supply Interruptions: Numerous articles report blackouts and load shedding, such as Zimbabwe's 18-hour outages or Zambia's eight-hour daily cuts. These power disruptions are often linked to aging infrastructure, drought-induced hydropower shortfalls, or fuel supply interruptions.
- Grid Collapse and Plant Failures: Systemic instability in national grids is a recurring issue, exemplified by Nigeria's grid collapses due to gas pipeline fires or mass shutdowns of power plants. These events often result in widespread service loss and affect both domestic and industrial users.
- Sector Investment Challenges: The power sector is frequently described as financially fragile, with reports highlighting liquidity crises, investor hesitance due to poor returns, and debt owed by government agencies to electricity distributors. These problems undermine capital expenditure on maintenance, metering, and network expansion.
- Regulatory and Legal Reform: Some governments are pursuing legal and institutional reform to attract investment or decentralize electricity generation (e.g., Lagos State's Power Sector Reform Law and embedded power projects). Such reforms seek to create an enabling environment for private sector participation.
- International Partnerships and Imports: In the face of domestic shortfalls, countries like Zimbabwe have turned to foreign electricity imports, sometimes after negotiating debt repayment terms, showing how regional interdependence is vital to power security.
- Renewable and Off-grid Initiatives: Media also covers alternative solutions for energy access, particularly in rural or unconnected areas. Off-grid solar power projects funded by organizations like the UNDP and BoI are presented as innovative and scalable responses to energy poverty.
- Environmental Factors and Natural Disasters: Electricity supply is also portrayed as vulnerable to external shocks, such as heavy rainfall causing blackouts in Nairobi or drought affecting hydroelectric output.

*Identification Prompt.* Africa power crisis, Africa Energy Crisis, Electricity supply crisis.

Transport (d30a)

*Summary Prompt.* This theme captures the development, state, and economic implications of transport infrastructure and systems, especially rail and road networks. Media coverage frames transport not just as a logistical necessity, but as a strategic asset tied to national development, regional trade integration, and industrialization.

Contextual Examples from News Coverage:

- Railway Infrastructure Expansion: A dominant theme is the investment in railway development, such as Nigeria's Wagon Assembly Plant, Ghana's Accra-Tema rail relaunch, and Ethiopia's industrial belt along the Ethio-Djibouti railway. These initiatives are framed as catalytic for economic transformation, boosting freight movement, industrial logistics, and commuter mobility.
- Regional Integration and Trade Facilitation: Projects like the Mombasa-Nairobi SGR and Addis Ababa-Djibouti electrified railway are frequently cited as examples of cross-border transport integration, especially supported by Chinese investment. These are expected to ease port access, reduce freight times, and enhance regional trade competitiveness.
- Urban and Suburban Connectivity: In countries like South Africa and Nigeria, expanding light rail systems (e.g., Gautrain or Ogun's Agbara-Berger line) are aimed at decongesting roads and improving urban commuting efficiency. Such investments are linked to improving worker productivity and access to economic hubs.
- Technology Transfer and Local Capacity Building: Transport infrastructure projects are also discussed in the context of technology localization. For example, Nigeria's partnership with Chinese contractors includes plans for rail manufacturing facilities and skill transfer to local engineers.
- Public-Private and Multilateral Financing: Articles highlight multi-stakeholder funding models, such as Uganda's expressway co-financed by the EU, AfDB, and AFD. This reflects a shift toward leveraging international capital for long-term infrastructure development.
- Restoration and Modernization Efforts: Government officials in several countries frame transport investments as part of a broader "restoration" of neglected systems, promising that rail transport will reclaim its role as a backbone of national economies.

*Identification Prompt.* African Railway Development, African Railway Development, Rail Infrastructures.

Customs and Trade (d30b)

*Summary Prompt.* This theme captures media discussions related to trade liberalization, customs procedures, regional economic integration, and institutional reforms aimed at facilitating the movement of goods across borders. It includes developments in tariff regimes, trade agreements, export promotion strategies, and efforts to modernize customs systems.

Contextual Examples from News Coverage:

- Continental Trade Integration (AfCFTA): The dominant media narrative centers on the African Continental Free Trade Area (AfCFTA), hailed as a historic agreement to remove 90% of tariffs across 55 African nations. Media emphasize its potential to boost intra-African trade, support industrialization, and reduce dependency on foreign markets.
- Customs Facilitation and Modernization: National efforts to ease cross-border trade are also covered, such as Rwanda's establishment of a National Trade Facilitation Committee (NTFC) in line with WTO commitments. These reforms aim to make customs procedures more efficient, transparent, and aligned with international standards.

- Export Development and Quality Standards: Some coverage focuses on improving the quality and competitiveness of exports (e.g., Nigeria's push to upgrade product standards under AfCFTA), suggesting that participation in trade pacts requires capacity building in production and logistics.
- Trade Promotion Events and Business Linkages: International trade shows (e.g., in Rajkot, India) and forums (e.g., the World Export Development Forum in Ethiopia) are framed as platforms for building bilateral ties, attracting investment, and facilitating business-to-business cooperation.
- Legal Instruments and Policy Negotiations: Several articles highlight the negotiation, ratification, and implementation processes of trade agreements. Discussions include divergent national interests, the need for harmonized rules, and legal infrastructure to support the free movement of goods.
- Concerns About Trade Imbalances: Not all media coverage is uncritical. Some articles reference fears among certain countries—such as Nigeria's initial hesitation—about the risk of market flooding by cheap imports, underscoring the complexities of liberalizing trade in uneven economic contexts.

*Identification Prompt* . African Free Trade, African economic integration, African economic integration.

Access to Finance (k30)

*Summary Prompt.* This theme encompasses the availability, affordability, and inclusiveness of financial services—particularly credit, loans, and investment capital—for individuals, businesses, and sectors critical to economic development. It includes both formal and informal financial mechanisms, regulatory reforms, financial literacy efforts, and targeted interventions by monetary authorities.

Contextual Examples from News Coverage:

- Credit to the Real Sector: Media frequently highlights the role of banks and the Central Bank in extending credit to agriculture, manufacturing, and MSMEs. This includes structured interventions such as the Commercial Agriculture Credit Scheme, Anchor Borrowers Programme, and Youth Entrepreneurship Development Programme, all designed to address lending gaps.
- Financial Inclusion: Articles emphasize ongoing efforts to bring underserved populations—especially youth and rural communities—into the financial system. These include financial literacy campaigns (e.g., in Ghana), mobile lending innovations, and policies focused on digital financial services to reach the unbanked.
- Interest Rate and Policy Debates: Regulatory debates feature prominently, such as the CBN's warning against excessive interest rates on intervention funds or the IMF's recommendation that Kenya remove interest rate caps. These highlight tensions between financial access, affordability, and market stability.
- Monetary and Development Policy Interventions: Media discusses the Central Bank's broader unconventional monetary policies that directly support job creation and lending to strategic sectors. Notably, over ₦1 trillion has been earmarked to stimulate the real economy, reflecting the financial sector's growing developmental role.
- Foreign Exchange Access and Trade Finance: Nigeria's Investors and Exporters (I&E) forex window, through which over $60 billion has been traded, is presented as a tool to facilitate smoother access to capital for exporters and importers, improving liquidity and investment climate.
- Recognition and Innovation in Lending: Institutions like Sterling Bank are profiled for their role in sector-specific finance, such as agriculture, while Fintech firms (e.g., FIDO in Ghana) are shown to be reshaping lending practices through technology and inclusive design.

*Identification Prompt.* Financial inclusion initiatives, Financial sector development, Financial sector development.

Access to Land (g30a)

*Summary Prompt.* This theme captures the availability, security, and equity of land ownership or usage rights, as well as the legal, political, and socio-economic barriers to acquiring, holding, or utilizing land. Media coverage includes issues such as land titling, land grabbing, historical dispossession, gender inequality in land access, and state-led land redistribution or reform programs.

Contextual Examples from News Coverage:

- Historical and Legal Land Injustice: Reports from Kenya and South Africa highlight land disputes rooted in colonial-era policies and discriminatory laws. Cases like the Waitiki farm evictions and the ANC-led land reform in South Africa demonstrate how unresolved historical grievances shape modern land politics and test legal systems.
- Lack of Formal Land Titles: A significant barrier to access is the lack of official documentation, with articles noting that up to 70% of Kenya's land remains untitled. This results in insecurity of tenure, vulnerability to land grabs, and constrained access to land-related credit or investment.
- Land Grabbing and Commercial Pressure: Media outlets report on land grabs by multinational corporations in countries like Nigeria, often under the guise of agricultural or infrastructure development. These stories emphasize the displacement of rural communities and raise concerns over transparency and accountability in land governance.
- Administrative and Policy Reforms: Some regions are modernizing land administration systems, as seen in Lagos State's push for digital land titling to reduce bureaucratic inefficiencies and support property rights. These reforms are often aimed at improving the ease of doing business and investment climate.

- Gender Inequality in Land Ownership: Media coverage highlights significant gender disparities—especially in rural Africa—where women provide the majority of agricultural labor but hold a fraction of land titles. Cultural norms and legal barriers are cited as key contributors to this persistent gap.
- Land Redistribution and Political Pledges: High-profile government initiatives, such as South Africa's mass land redistribution plans, are framed as politically consequential, economically necessary, and socially contentious. Leaders describe land access as central to justice, economic empowerment, and national development.
- Agricultural Land Use and Commercialization: Broader access to land is also discussed in relation to rural transformation, such as in Rwanda's encouragement of smallholder farmers to move from subsistence to commercial agriculture through improved land access and irrigation.

*Identification Prompt.* Land ownership conflicts, Land rights reform, Land ownership struggles.

Crime, Theft, and Disorder (i30)

*Summary Prompt.* This theme captures the occurrence, reporting, and institutional response to unlawful acts that threaten personal safety, property rights, and public order. It includes violent crime (e.g. armed robbery, kidnapping), organized or culturally embedded practices (e.g. cattle rustling), policing and security sector issues, as well as broader social conditions contributing to criminality and disorder.

Contextual Examples from News Coverage:

- Violent and Organized Crime: Media frequently reports on criminal networks involved in armed robbery, kidnapping for ransom, cattle rustling, and gunrunning. Parades of arrested suspects with weapons in Nigeria or raids on ex-militant groups illustrate both the scale of the threat and the response mechanisms from state authorities.
- Cattle Rustling as Socio-Cultural Crime: In Kenya's Nyakach region and other parts of East Africa, cattle rustling is presented not only as theft but also as a culturally embedded form of violent contestation that has evolved into a major security concern.
- Militancy and Weaponized Disorder: Articles detail how ex-militants and militiamen contribute to rising insecurity—particularly when disarmament or reintegration efforts fail. This includes criminal activity by disenfranchised former combatants, especially following the loss of government contracts or protection roles.
- Human Trafficking and Child Exploitation: Reports from Uganda raise concerns about child trafficking under the guise of foreign adoptions, pointing to regulatory gaps and lack of oversight in vulnerable systems, further indicating how disorder intersects with international crime.
- Security Infrastructure and Military Protection: In response to rising threats, some governments (e.g., Ghana) are developing strategic military bases to secure critical national infrastructure (e.g., oil and gas installations), highlighting efforts to protect economic assets from criminal activity.
- Police and Legal System Challenges: Media also critiques policing practices, including the public parading of suspects, which raises concerns about due process and institutional impunity. This is often presented as a symptom of deeper systemic dysfunction in criminal justice administration.
- Conflict-Related Instability: Violent inter-communal conflict and political disorder are highlighted as major contributors to human insecurity, displacement, and slowed development (e.g., UNDP reports on violent deaths and mass displacement in Kenya).

*Identification Prompt.* Insecurity & Instability, Rising insecurity crisis, Insecurity & Instability.

**Calculation of the Semantic Brand Score (SBS)**

This part of the appendix details the calculation of the Semantic Brand Score, originally introduced by Fronzetti Colladon (2018). As discussed in Section 3.4 (*Measuring Semantic Importance*), the SBS quantifies the semantic importance of a word or concept/theme, defined as a set of words, within large text corpora. The SBS is a composite indicator that integrates three dimensions derived from text mining and social network analysis: *prevalence*, *diversity*, and *connectivity*. In our application, the SBS is calculated separately for each region-year corpus, using only the news articles assigned to that specific region and year. The same preprocessing and network-construction procedure is applied to all region-year corpora. Each of the 13 themes is represented by a predefined set of terms, which are treated jointly as a single conceptual unit in the SBS calculation.

Before calculating the SBS, the text corpus undergoes a series of preprocessing steps. These include removing punctuation, stop-words, and special characters; converting all text to lowercase; and stemming words using the Porter algorithm (Porter, 2006) to reduce them to their base form. These steps aim to reduce linguistic complexity while preserving the semantic content of the text. Preprocessing is performed at the article level, so that word co-occurrences are identified within articles and never across article boundaries.

Once preprocessed, the text is transformed into a co-occurrence network. In this network, each node represents a unique word, and edges are weighted by the frequency with which pairs of words co-occur within a given context window. The resulting co-occurrence network is symmetrized, self-loops are removed, and repeated co-occurrences between the same pair of words are represented by a single undirected edge whose weight is equal to their co-occurrence frequency. An example of such a semantic network is shown in Figure A4.

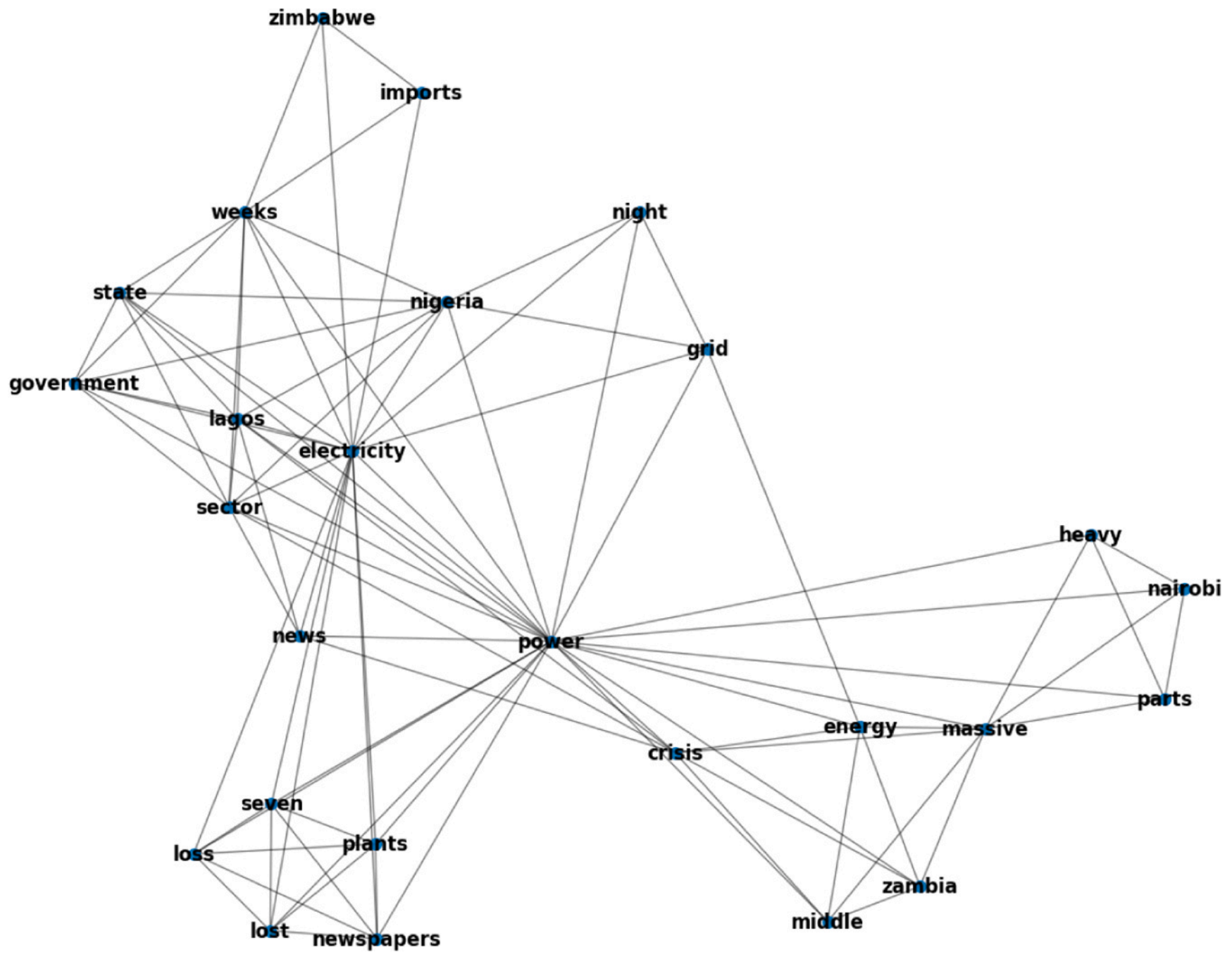

**Fig. A4.** Example of a Semantic Network

Representing text as a network has two primary advantages. First, it captures the relationships among words, highlighting contextual linkages. Second, it allows the application of social network analysis techniques to textual data.

SBS Components

*Prevalence* captures the frequency with which a concept (i.e., a set of words) appears in the corpus. It reflects how often a theme is mentioned and, by extension, how salient or recognizable it may be to readers (Keller, 1993). It is calculated as the total frequency of the terms associated with the concept.

*Diversity* assesses the uniqueness and diversity of the semantic contexts in which a concept occurs. It is based on the distinctiveness centrality metric introduced by Fronzetti Colladon and Naldi (2020), which evaluates the heterogeneity and uniqueness of a node's connections in the network. Specifically, in a network with n nodes, the distinctiveness of node i is calculated as:

$$D(i) = \sum_{\substack{j=1 \\ j \neq i}}^{n} \log_{10}\frac{n-1}{\deg(j)} I_{\{w_{ij}>0\}}$$

where *deg*(*j*) is the degree of node *j*, and $I_{\{w_{ij}>0\}}$ is an indicator function equal to 1 when the edge exists and has positive weight and equal to zero otherwise.

*Connectivity* reflects how well-integrated a concept is within the overall discourse, indicating its brokerage power. It is measured using betweenness centrality (Brandes, 2001a; Freeman, 1978), which captures the extent to which a node (or cluster of nodes) lies on the shortest paths between other nodes in the network:

$$C(i) = \sum_{j \neq k \neq i} \frac{\sigma_{jk}(i)}{\sigma_{jk}}$$

Where $\sigma_{jk}$ is the total number of shortest paths from node *j* to node *k*, and $\sigma_{jk}(i)$ is the number of those paths that include node *i*.

The final SBS is obtained by summing the normalized values of the three components: *prevalence*, *diversity*, and *connectivity*. Each component is scaled using min-max normalization:

$$SBS = \text{norm(Prevalence)} + \text{norm(Diversity)} + \text{norm(Connectivity)}$$

This normalization ensures comparability of SBS values across different periods by mapping the scores onto a consistent scale ranging from 0 to 3.

## Data availability

The authors do not have permission to share news data. All numerical data used in the paper will be made available upon reasonable request to the second author.